\documentclass{article}
\usepackage{iclr2027_conference,times}
\usepackage[utf8]{inputenc}
\usepackage[T1]{fontenc}
\usepackage{amsmath,amsfonts,amssymb}
\usepackage{graphicx}
\usepackage{placeins}
\usepackage{booktabs}
\usepackage{microtype}
\usepackage{xcolor}
\usepackage{xurl}
\usepackage[hidelinks]{hyperref}
\graphicspath{{figures/}}

\hypersetup{
  pdftitle={Decoding Register Tokens Preserves Global Scene Details},
  pdfauthor={David Drexlin, Julius Hense, Jonas Dippel},
  pdfsubject={Preprint. Under review.},
  pdfdisplaydoctitle=true
}

\title{Decoding Register Tokens\\
Preserves Global Scene Details}

\author{%
  \begin{tabular}{@{}c@{}}
    {\fontsize{11}{13}\selectfont\bfseries
      David Jacob Drexlin\textsuperscript{1,2,*}\quad
      Julius Hense\textsuperscript{1,2}\quad
      Jonas Dippel\textsuperscript{3}}\\[0.65em]
    \textsuperscript{1}Machine Learning Group, Technische Universit\"at Berlin\\
    \textsuperscript{2}BIFOLD -- Berlin Institute for the Foundations of Learning and Data\\
    Franklinstra{\ss}e~28/29, 10587 Berlin, Germany\\[0.35em]
    \textsuperscript{3}Aignostics, Germany\\[0.6em]
    \textsuperscript{*}Corresponding author:
    \href{mailto:drexlin@tu-berlin.de}{\texttt{drexlin@tu-berlin.de}}
  \end{tabular}%
}
\date{}

\makeatletter
\renewcommand{\@maketitle}{%
  \vbox{%
    \hsize=\textwidth
    \linewidth=\hsize
    \parindent=0pt
    \parskip=0pt
    \centering
    {\normalfont\fontsize{18}{21}\selectfont\bfseries\@title\par}%
    \vspace{0.9em}%
    {\normalfont\fontsize{10}{12}\selectfont\@author\par}%
    \vspace{0.4em}%
  }%
}
\makeatother

\iclrfinalcopy

\begin{document}
\maketitle
\lhead{Preprint. Under review.}

\begin{abstract}
Pretrained vision transformers learn structured image representations
spanning spatial patch tokens and non-spatial tokens. Recent analyses
suggest that non-spatial register tokens encode global scene properties such as
illumination, blur, and style. Yet patch-based image decoders discard
these outputs, leaving the decoder to recover image appearance from
patches alone. We investigate whether retaining register features
improves decoding when the complete spatial patch representation is
already available.
We study this question using representation autoencoders (RAEs) with
frozen register-equipped encoders. Our decoder processes four register
tokens together with all patch tokens through joint self-attention.
Appearance probes, paired reconstruction-error analysis, and fixed-decoder
register substitutions examine which global image properties are
accessible from registers and how their values affect reconstructed
pixels. For generation, we train a multi-head attention pooling
(MHAP) predictor to recover register features from sampled patches,
keeping the pretrained diffusion model and its samples unchanged.
On ImageNet-1K, our ViT-B decoders achieve mean reconstruction FIDs 
of 0.472 with DINOv2 and 0.442 with DINOv3, reductions of 9.5\% and 5.2\%
relative to their respective patch-only baselines. The corresponding PSNRs are 19.73 
and 20.54\,dB. For generation, our DINOv2 models achieve a mean FID of 1.641, 9.8\% below
the patch-only baseline on identical sampled patches. 

\end{abstract}

\section{Introduction}
\label{sec:introduction}

Representation autoencoders (RAEs) connect pretrained visual
representations with latent image generation through a two-stage
pipeline~\citep{zheng2025rae,tong2026scalerae}. A decoder learns to
reconstruct images from a frozen encoder, and a diffusion transformer
(DiT) learns to generate features in that space. At inference,
sampled features are decoded into pixels. The decoder must therefore
use a representation shaped by the encoder's pretraining objective,
rather than learned jointly for reconstruction.

Patch-based RAEs retain spatial patch tokens while discarding
non-spatial encoder outputs. For register-equipped DINOv2 and
DINOv3 encoders, this excludes four register features already
computed in the same forward
pass~\citep{oquab2023dinov2,darcet2023registers,simeoni2025dinov3}.
Registers support non-local computation and reduce artifacts in
spatial features~\citep{darcet2023registers}. More recently,
\emph{Into the Rabbit Hull} identified register-specialized concepts
associated with illumination, blur, and style~\citep{fel2025rabbithull}.
Such properties matter to reconstruction: recovering recognizable
objects does not ensure that their lighting, color relationships,
or surrounding textures match the source image.

Discarding registers leaves the decoder to recover these properties
from patches alone. Self-attention allows aggregation across all
patches, but recoverability does not guarantee that a decoder learns
an equally effective readout under its reconstruction objective and
training budget. Image-reconstruction losses supervise the output,
not recovery of the encoder's register state. Registers may therefore
make appearance information easier to use even when related
information remains in patches. Our question is:
\emph{does supplying register features improve image decoding when
the complete spatial patch representation is retained?}

We prepend four register outputs to the patch sequence and decode
both through joint self-attention, keeping the encoder frozen and
introducing no additional decoder loss. On ImageNet-1K, DINOv2 and
DINOv3 ViT-B decoders achieve 9.5\% and 5.2\% lower reconstruction
FID than their respective patch-only references. 
Without additional training, register-input decoding also improves
rFID, PSNR, and DISTS on NINCO and Places365 for both encoder families.
To examine the image properties involved, we combine appearance probes with a
separate paired ImageNet-100 reconstruction analysis. The latter
finds lower error in both mean RGB color and spatially centered
images: the benefit extends beyond correcting a constant color
offset. Fixed-decoder substitutions and interpolations further
show that register values influence color and blur-related
appearance while receiver patches remain unchanged. Together,
these experiments examine both reconstruction fidelity and the
decoder's response to changes in register state.

For generation, pretrained patch-latent DiTs do not produce
register features. We train a multi-head attention pooling (MHAP)
predictor~\citep{lee2019settransformer} independently of the image
decoder to recover registers from sampled patches.
Within a fixed decoder, predicted registers outperform mean and
shuffled replacements on encoded images, although encoder-produced
registers retain the lowest reconstruction errors.
Decoding identical DINOv2 patch samples with recovered registers
achieves mean generation FID 1.641, compared with 1.819 for the
patch-only reference, a 9.8\% reduction.
This improves the decoding of an existing generator's samples
without retraining the DiT or adding register tokens to its
diffusion process.

\section{Related Work}
\label{sec:related-work}

\paragraph{Representation autoencoders and generative tokenizers.}
Latent diffusion models and diffusion transformers synthesize images
in learned feature spaces~\citep{rombach2022ldm,peebles2023dit}.
RAEs replace reconstruction-trained encoders with frozen vision
backbones and learn decoders over their features~\citep{zheng2025rae}.
Scale-RAE extends this approach to text-to-image
generation~\citep{tong2026scalerae}, while RAEv2 aggregates
features from multiple encoder layers and combines RAE latents
with intermediate diffusion-feature alignment~\citep{singh2026raev2}.
LV-RAE adds a learned local-variation encoder to improve
reconstruction~\citep{liu2026lvrae}. We retain all spatial patches
and expose four register outputs from the same encoder, rather
than adding another image encoder or modifying the representation
modeled by diffusion.

\paragraph{Decoding pretrained representations.}
Image synthesis provides a way to examine which image properties
a representation preserves. Diffusion Autoencoders learn semantic
codes together with a diffusion decoder~\citep{preechakul2022diffae}.
RCDM visualizes self-supervised representations through conditional
generation~\citep{bordes2022rcdm}, and unCLIP generates images
from CLIP image embeddings~\citep{ramesh2022unclip}.
Pretrained representations also guide diffusion training through
feature alignment and architectural changes, as in REPA and
DDT~\citep{yu2025repa,wang2025ddt}. Our register-recovery pathway
instead operates after sampling, leaving pretrained diffusion
weights and sampled patches unchanged.

\paragraph{Registers and global appearance.}
\citet{darcet2023registers} introduced register tokens to reduce
artifacts associated with non-local computation in vision-transformer
patch features. Subsequent work examines whether registers require
explicit training and how their role varies across
architectures~\citep{jiang2025testtimeregisters,baxevanakis2026reassessregisters}.
\emph{Into the Rabbit Hull} analyzes concept organization in DINO
representations and identifies register-specialized concepts
associated with illumination, blur, and style~\citep{fel2025rabbithull}.
We study the consequences for reconstruction and appearance control
by using register outputs as decoder inputs, without
altering register computation inside the encoder.

\section{Method}
\label{sec:method}

We extend patch-based RAEs~\citep{zheng2025rae} to decode encoder
registers alongside spatial patches, and recover registers from
sampled patches for generation.

\subsection{Encoder representations and decoder inputs}
\label{sec:problem-setup}

For an image $x$, a frozen register-equipped DINOv2 or DINOv3
encoder produces $N$ patch features
$Z_p(x)\in\mathbb{R}^{N\times d}$ and four register outputs
$Z_r(x)\in\mathbb{R}^{4\times d}$, where $d$ is the feature
dimension. Both decoding conditions use the same encoder and
complete patch sequence; only the patch-and-register decoder
receives the register outputs.

Let $D_P$ and $D_R$ denote the patch-only and register-input
image decoders, with inputs expressed in encoder-feature
coordinates. Their reconstructions are
\begin{equation}
\begin{aligned}
\hat x_P &= D_P\!\left(Z_p(x)\right),\\
\hat x_R &= D_R\!\left(Z_p(x),Z_r(x)\right).
\end{aligned}
\label{eq:decoder-interfaces}
\end{equation}

\begin{figure}[t]
\centering
\includegraphics[width=0.95\linewidth]{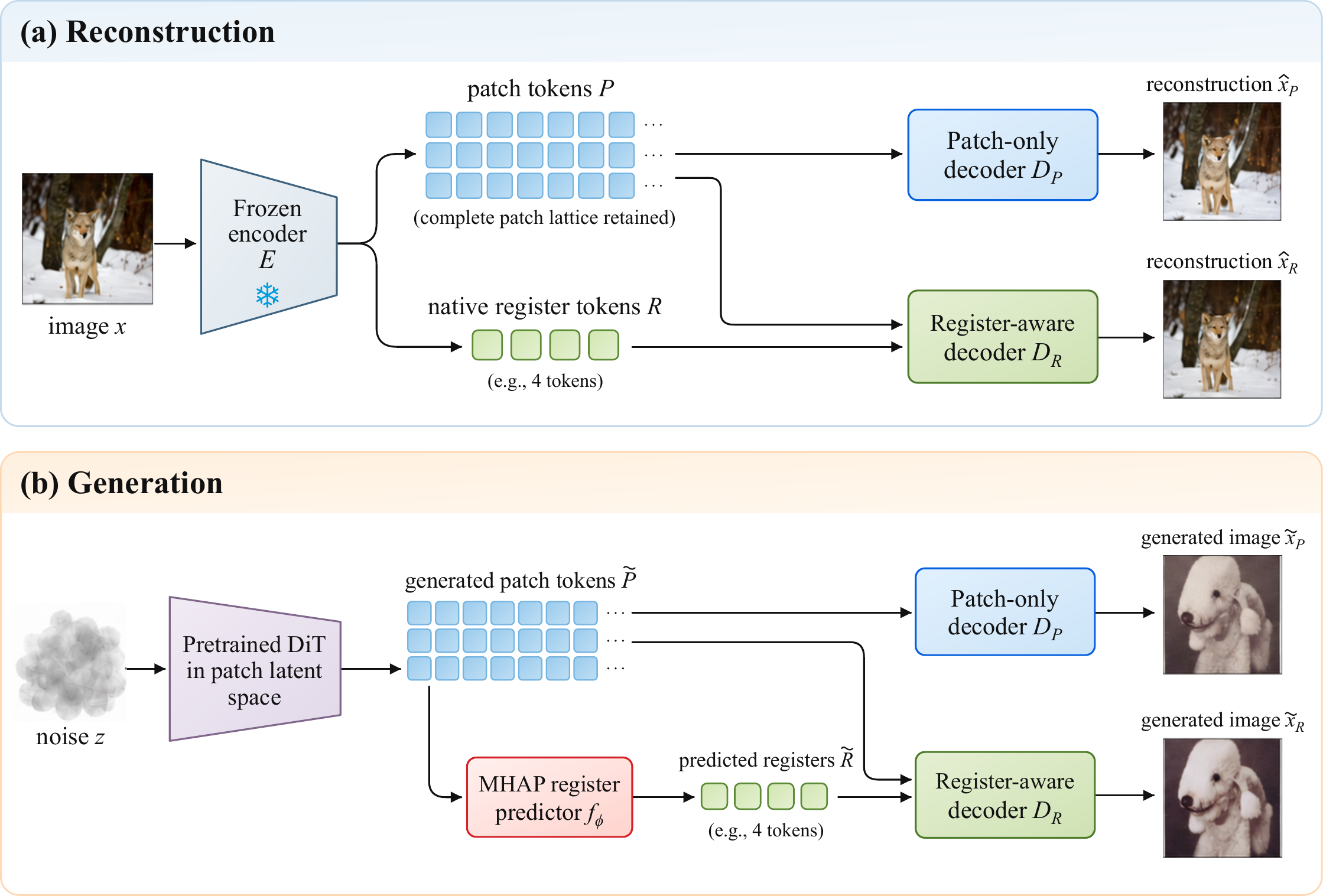}
\caption{\textbf{Decoding patches and registers.}
(a) The patch-only decoder $D_P$ receives all spatial patches;
$D_R$ additionally receives four encoder registers.
(b) MHAP recovers registers from sampled patches before decoding
with $D_R$. Both generation routes use the same pretrained DiT
and sampled patches.}
\label{fig:overview}
\end{figure}

Figure~\ref{fig:overview} summarizes reconstruction and generation.

\subsection{Decoding patches and registers}
\label{sec:native-register-decoding}

Each decoder learns token-wise input projections. We denote the
patch and register projections by $W_p$ and $W_r$, respectively.
The register decoder prepends the projected registers:
\begin{equation}
H^0=
\bigl[W_r Z_r+b_r;\,W_p Z_p+b_p\bigr].
\label{eq:prepend}
\end{equation}
The semicolon denotes token concatenation.
The decoder's own learned class token and positional embeddings
are omitted from this expression; their conventions are specified
in Appendix~\ref{appendix:feature-loss-details}.

Both decoder variants use a ViT-B transformer backbone and
RGB reconstruction, perceptual, Gram, and adversarial image losses.
The register-input decoder additionally receives four
encoder registers through joint self-attention.
For further details on decoder training
we refer the reader to Appendix~\ref{appendix:feature-loss-details}.
No additional register-prediction loss is used in decoder training.
Section~\ref{sec:register-interventions} fixes a trained decoder
and its patches to test changes in register values.

\subsection{Register recovery for generation}
\label{sec:register-recovery}

A pretrained patch-latent DiT does not produce registers.
We recover them with a multi-head attention pooling (MHAP)
predictor~\citep{lee2019settransformer}: four learned queries
attend to all patches and predict the four registers in
encoder-defined order. Training uses image-matched patch--register
pairs from the frozen encoder and mean squared error in standardized
register coordinates, independently of the image decoder and
diffusion model. The loss and coordinate conversions are specified
in Appendix~\ref{appendix:posthoc-register-prediction}.

Let $\widetilde Z_p$ denote sampled patches in encoder-feature
coordinates and $f_\phi$ include conversion of predicted registers
back to those coordinates. Generation uses
\begin{equation}
\widehat Z_r=f_\phi(\widetilde Z_p),\qquad
\widetilde x_R=D_R(\widetilde Z_p,\widehat Z_r).
\label{eq:recovered-generation}
\end{equation}

The patch-only reference decodes the same sampled patches with
$D_P$. The register decoder and coordinate conversions remain
fixed when native registers are replaced by predictions.
Recovering registers once after sampling provides decoder features
without additional diffusion targets; a separate joint-generation
experiment appears in
Appendix~\ref{appendix:joint-auxiliary-generation}.

\section{Experiments}
\label{sec:experiments}

We test appearance information accessible from registers, their effect
on reconstruction, and their recovery for generation. ImageNet-100
supports probes and paired appearance analysis; ImageNet-1K provides
the principal reconstruction and generation evaluations.

\paragraph{Setup and evaluation.}
We use 12-block ViT-B decoders at $256\times256$ resolution;
ImageNet-1K register decoders are trained for 16 epochs.
Reconstruction and generation FID (rFID/gFID) compare
Inception-feature distributions~\citep{heusel2017gans}.
Preserving global scene details also requires fidelity to
the source image's appearance, including color and brightness.
For reconstruction, we therefore report PSNR for paired pixel
agreement and DISTS for paired structure-and-texture
discrepancy~\citep{ding2021dists}; higher PSNR and lower DISTS
indicate closer reconstruction.
Wavelet-distance results, metric protocols, and uncertainty
definitions appear in
Appendices~\ref{appendix:appearance-metric-sensitivity}
and~\ref{appendix:evaluation-uncertainty}.

\subsection{Register state and global appearance}
\label{sec:probe-complementarity}
Ridge-regression probes predict an $8\times8$ RGB thumbnail,
a color histogram, and low-frequency discrete cosine transform
(DCT) coefficients from frozen DINOv2 features. The zero-frequency
(DC) coefficient is proportional to each channel's spatial mean;
the selected nonzero-frequency (AC) coefficients describe
low-frequency spatial variation. We compare mean-pooled patches
$p=\operatorname{mean}(Z_p)$ with the same features augmented
by the four-register bank $r=\operatorname{vec}(Z_r)$.
For each target $y$, we measure the change in held-out
coefficient of determination, $R^2$:

\begin{equation}
\Delta R^2(r\mid p)
=R^2([p;r]\rightarrow y)-R^2(p\rightarrow y).
\label{eq:probe-increment}
\end{equation}
Feature standardization and regularization selection use only
the training portion of each split.
Table~\ref{tab:conditional-complementarity-main} reports
held-out prediction scores.

\begin{table}[t]
\centering
\small
\begin{tabular}{lrrrr}
\toprule
Features & Thumbnail & Color hist. & DCT DC & DCT AC\\
\midrule
Patch mean & 0.211 & 0.288 & 0.735 & 0.038\\
Patch mean + registers & 0.228 & 0.296 & 0.820 & 0.068\\
\bottomrule
\end{tabular}
\caption{\textbf{ImageNet-100 appearance probes.}
Held-out $R^2$. DC measures spatial mean; AC contains nonzero
low-frequency DCT coefficients.}
\label{tab:conditional-complementarity-main}
\end{table}

Adding registers improves all four prediction targets, with the
largest increase for mean RGB color: DC $R^2$ rises from
0.735 to 0.820.

To test whether the benefit persists without averaging away patch
positions, a separate probe retains the complete flattened patch
sequence. Across three shared outer splits, adding native registers
increases mean-color (DC) $R^2$ from 0.573 to 0.663 and thumbnail
$R^2$ from 0.241 to 0.247. Aggregate non-DC prediction changes
little ($\Delta R^2=-1.9\times10^{-4}$).
With identical splits and effective regularization candidates,
zero-register inputs reproduce the patch-only predictions exactly;
shuffled and Gaussian ablation experiments provide no average gain on the
DC or thumbnail targets
(Appendix~\ref{appendix:full-patch-probe}).

These results show that registers expose global color in a form
that the ridge readout can exploit alongside the patches.
They characterize accessibility to this readout, rather than
the total information available in the patch representation.

\begin{figure}[t]
\centering
\includegraphics[width=\linewidth]{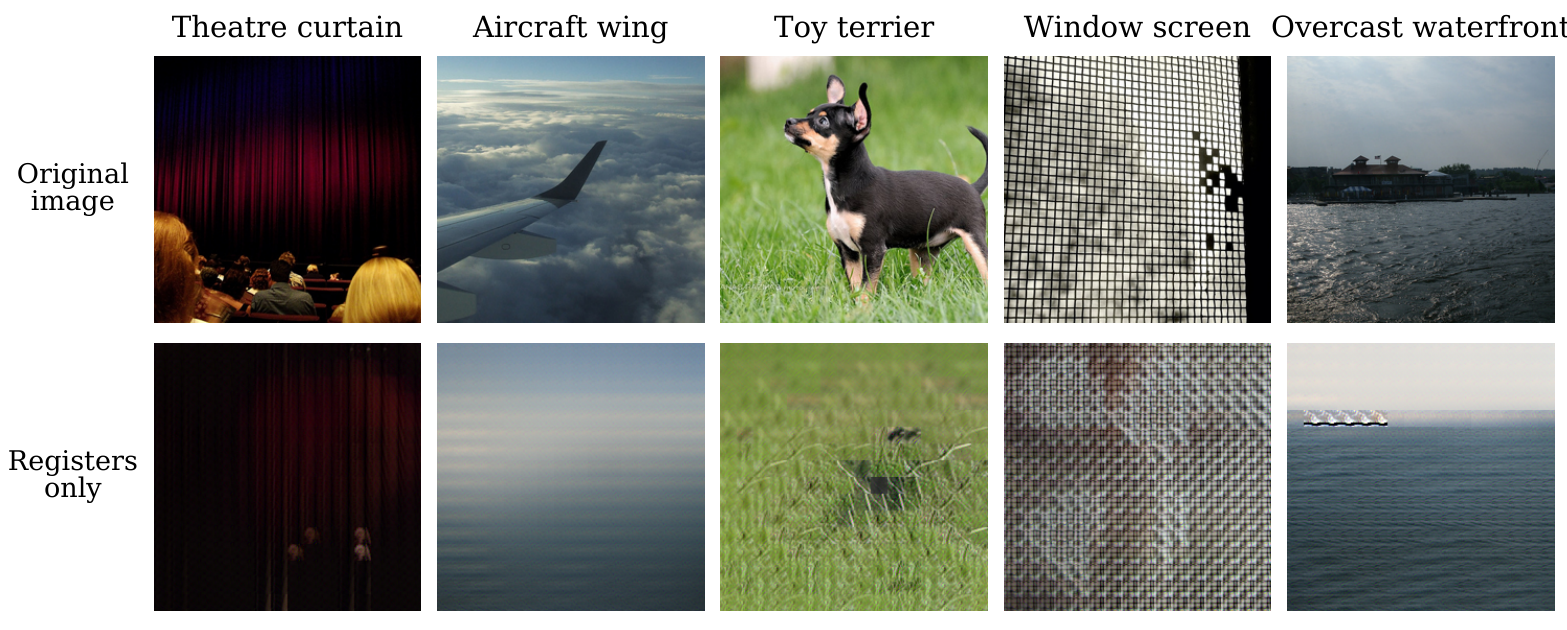}
\caption{\textbf{Reconstruction from four registers alone.}
Originals (top) and reconstructions from four registers without patches
(bottom). Color and texture are partly recovered, while substantial
object and spatial detail is missing. Decoder and selection details
appear in Appendix~\ref{appendix:register-only-reconstruction}.}
\label{fig:register-only-original-comparison}
\end{figure}

A decoder receiving only the four registers provides an
image-space illustration: it recovers aspects of color and texture while losing substantial
object and spatial detail
(Figure~\ref{fig:register-only-original-comparison}).
This behavior is consistent with the register specialization
identified through dictionary learning in
\emph{Into the Rabbit Hull}, which associates register-specific
concepts with illumination, motion blur, and
style~\citep{fel2025rabbithull}.

\subsection{Where register inputs improve reconstruction}
\label{sec:appearance-effects}
\label{sec:global-recon}

We compare patch-only and register-input DINOv2 reconstructions
on the same 5,000 ImageNet-100 validation images for each of three
decoder seeds. Register inputs increase PSNR by 0.473\,dB 
and reduce DISTS by 0.007 relative to patch-only decoding,
with paired-seed 95\% intervals $[0.464,0.481]$\,dB and $[-0.008,-0.006]$, respectively.
Both weighted DISTS components also decrease in every seed
(Appendix~\ref{appendix:dists-components}).

\paragraph{Mean color and spatial variation.}
To locate the pixel-error reduction, we decompose RGB MSE into
a channel-mean term and a spatially varying term.
Let $\mu(x)$ denote the vector of channel means and
$x^c=x-\mu(x)$. Then
\begin{equation}
\operatorname{MSE}(x,\hat x)
=\frac{1}{3}\left\|\mu(x)-\mu(\hat x)\right\|_2^2
+\operatorname{MSE}(x^c,\hat x^c).
\label{eq:appearance-centered-mse}
\end{equation}
The first term measures errors in average RGB color.
The second measures errors after centering the source and
reconstruction separately by their own channel means.
Centered values remain floating point and are not clipped.

Mean RGB MSE decreases by 8.52\%.
Of the absolute reduction, 43.21\% occurs in the channel-mean
term and 56.79\% in the centered term.
Centered MSE decreases by 5.28\% on average, although it increases
in 21.63\% of image--seed pairs.
The average benefit therefore extends beyond correcting
image-wide color offsets to improving spatially varying
reconstruction, with adverse cases retained in the analysis.

Residual-spectrum and wavelet-distance measurements appear in
Appendix~\ref{appendix:appearance-analysis}.
Figure~\ref{fig:reconstruction-generation}(a) and
Appendix~\ref{appendix:qualitative-reconstruction} provide
illustrative signed reconstruction-error maps for the
ImageNet-1K models.

\paragraph{Decoder ablations.}
Registers yield lower rFID than CLS alone in both input studies;
adding CLS to the register bank gives little further improvement
(Appendix~\ref{appendix:input-choice}).
A separate CLS-plus-register comparison retains an rFID advantage
over patch-only decoding across ViT-B, ViT-L, and ViT-XL
(Appendix~\ref{appendix:decoder-capacity}).
Learned-token and patch-summary ablations appear in
Appendix~\ref{appendix:summary-controls}.
On ImageNet-1K, native-register reconstruction and MHAP-based
generation achieve lower FID than the two tested internal
register-supervision variants
(Appendix~\ref{appendix:register-supervision}).

\subsection{Fixed-decoder register interventions}
\label{sec:register-interventions}

The reconstruction analysis identifies improvements in both
mean color and spatial variation. We next examine how a trained
decoder responds to the register values themselves:
which aspects of reconstructed appearance can change when
the complete patch representation remains fixed?

We first use a controlled appearance transformation.
For each image, we extract features from the original and from
a version modified by a combined color-and-saturation transformation
$T$, obtaining $(P,R)$ and $(P_T,R_T)$.
With decoder weights fixed, we evaluate all four combinations:
\begin{equation}
D_R(P,R),\quad D_R(P,R_T),\quad
D_R(P_T,R),\quad D_R(P_T,R_T).
\label{eq:appearance-four-way}
\end{equation}
This compares native and substituted registers with either
the original patches $P$ or the transformed patches $P_T$
held fixed. Register values are substituted in encoder-feature
coordinates; normalization, token positions, and output
processing remain unchanged.

With original patches fixed, substituting the transformed
registers moves reconstructed mean RGB toward the transformed
image. The mean directional coefficient is 0.499, with a
paired-seed 95\% interval of $[0.475,0.524]$.
This statistic measures the reconstruction's mean-color response
along the applied transformation direction; it does not
attribute a fraction of the patch-versus-register reconstruction
gain to register values.

The response also extends beyond that direction.
The orthogonal RGB-change norm is 0.0155, and classifier-logit
cosine similarity to the original image decreases from
0.929 to 0.924.
Thus, the decoder uses register values to alter reconstructed
appearance, with accompanying off-target changes.
The transformation and response statistics are defined in
Appendix~\ref{appendix:appearance-interventions}.

\paragraph{Register interpolation.}
The controlled substitution transfers registers between two
versions of the same image. We next explore what happens when
the donor or register prompt is a different scene, tracing the reconstruction
between the receiver's native register state and the donor's state.

\begin{figure}[t]
\centering
\includegraphics[width=\linewidth]
{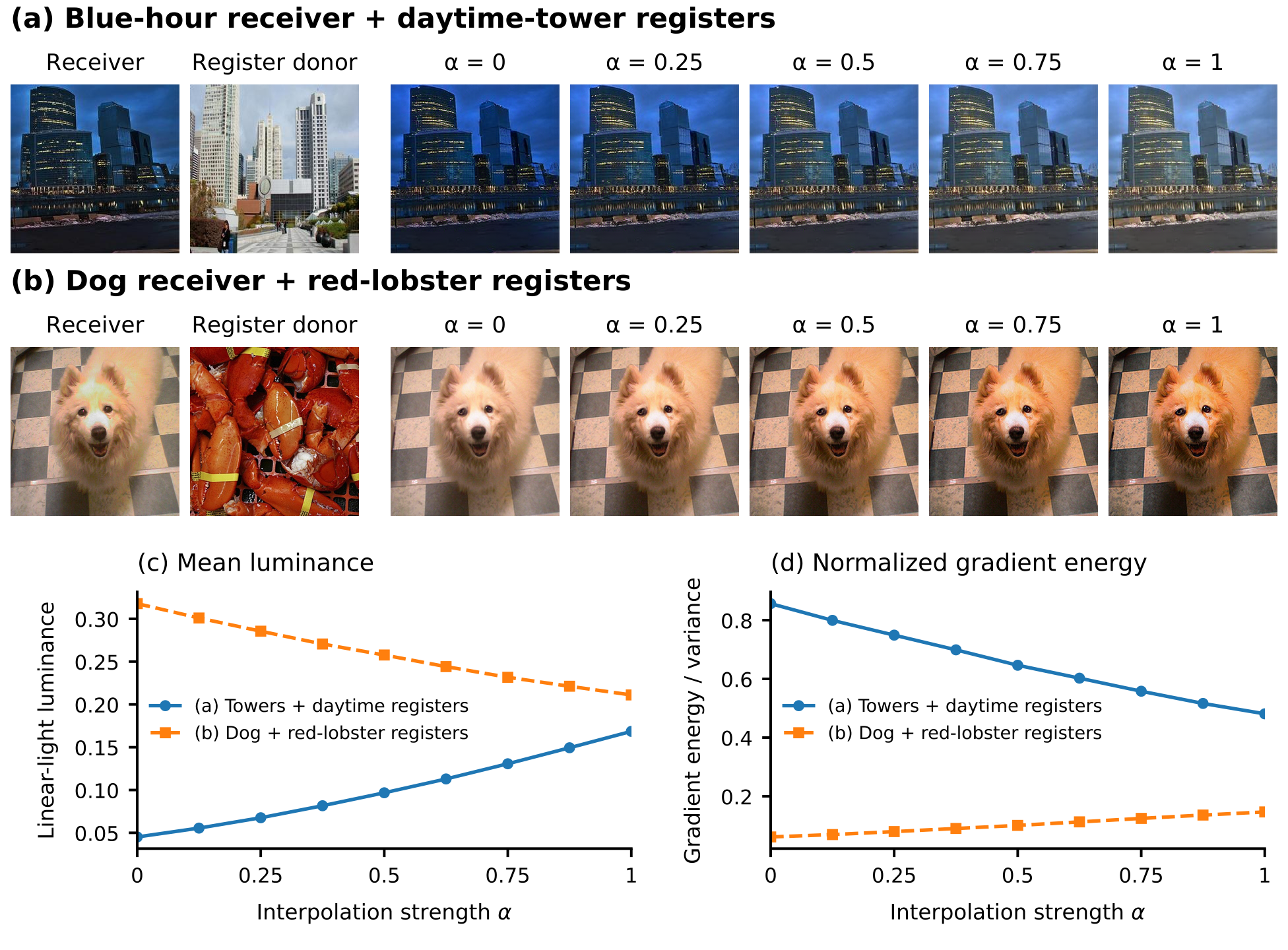}
\caption{\textbf{Appearance changes through register interpolation.}
(a) A blue-hour city receiver with registers from daytime towers.
(b) A dog receiver with registers from a red-lobster image.
Within each row, the receiver supplies the same complete patch
sequence throughout. Each of the four register slots is
interpolated separately toward its corresponding donor slot,
using a shared interpolation coefficient $\alpha$.
Decoder weights and token positions remain fixed.
At $\alpha=0$, decoding uses the receiver's native registers;
at $\alpha=1$, it uses the donor's registers.
Five of nine evaluated interpolation values are displayed.
(c,d) Mean linear-light luminance and variance-normalized gradient
energy for the two individual trajectories.
Both rows use an ImageNet-1K-trained DINOv2 ViT-B decoder.}
\label{fig:register-appearance-control}
\end{figure}

For receiver registers $R$ and donor registers $R^\star$,
\begin{equation}
R_\alpha=\operatorname{SLERP}(R,R^\star;\alpha),\qquad
\hat x_\alpha=D_R(P,R_\alpha),\quad \alpha\in[0,1].
\label{eq:register-appearance-control}
\end{equation}
Corresponding slots share $\alpha$; receiver patches, decoder
weights, and positions remain fixed.
Figure~\ref{fig:register-appearance-control} shows two of
20 retained directed trajectories. Across all 20, endpoint
changes follow the donor-defined direction in 19/20 cases for
luminance and 20/20 for normalized gradient energy; 14/20 and
17/20 trajectories, respectively, are monotonic over the nine
sampled values. Direction is defined by the donor's versus
receiver's native reconstruction
(Appendix~\ref{appendix:register-interpolation}).

\subsection{ImageNet-1K reconstruction}
\label{sec:in1k-reconstruction}

We next evaluate reconstruction at ImageNet-1K scale with
both DINOv2 and DINOv3.
Each decoder reconstructs the same 50,000 validation images
(Table~\ref{tab:in1k-stage1}).
Register-input models achieve 9.47\% lower mean rFID with
DINOv2 and 5.24\% lower with DINOv3.
For each encoder, all three register-input runs score below
all three patch-only runs.

\begin{table}[t]
\centering
\small
\begin{tabular}{llcc}
\toprule
Encoder & Decoder input & rFID $\downarrow$
        & PSNR (dB) $\uparrow$\\
\midrule
DINOv2
& Patches
& $0.5211\pm0.0043$
& $19.4901\pm0.0072$\\
& Patches + registers
& $\mathbf{0.4717}\pm0.0032$
& $\mathbf{19.7262}\pm0.0012$\\
\midrule
DINOv3
& Patches
& $0.4669\pm0.0026$
& $20.2994\pm0.0182$\\
& Patches + registers
& $\mathbf{0.4424}\pm0.0014$
& $\mathbf{20.5406}\pm0.0156$\\
\bottomrule
\end{tabular}
\caption{\textbf{ImageNet-1K reconstruction.}
Mean $\pm$ sample standard deviation across three ViT-B decoder
seeds. rFID measures distributional similarity; PSNR measures
pixel agreement with the source image.}
\label{tab:in1k-stage1}
\end{table}

The distributional improvements are accompanied by better
pixel agreement with the source images.
Register-input decoders achieve mean PSNRs of 19.73 and 20.54\,dB
for DINOv2 and DINOv3, respectively, and reduce mean RGB
reconstruction MSE by 3.28\% and 2.96\% relative to their
patch-only references. MSE is lower for every decoder seed
for both encoder families.
The corresponding DISTS scores are 0.1044 and 0.0967;
complete paired-fidelity measurements appear in
Table~\ref{tab:in1k-register-fidelity} in the appendix.

Figure~\ref{fig:reconstruction-generation}(a) illustrates
the DINOv2 reconstruction differences, with signed error maps
showing where register inputs improve or worsen pixel agreement.
Additional examples appear in
Figure~\ref{fig:appendix-in1k-reconstruction-maps}.
The same ImageNet-1K-trained decoders also transfer without retraining
to NINCO and Places365
(Appendix~\ref{appendix:ood-reconstruction}).
In the seed-0 evaluation, register inputs reduce rFID by
6.18\%/43.93\% for DINOv2 and 5.18\%/28.14\% for DINOv3
on NINCO/Places365, respectively.
These distributional gains coincide with improved paired fidelity:
PSNR increases by 0.17--0.40\,dB and DISTS decreases by
0.0049--0.0065 across the four comparisons.

\begin{figure}[t]
\centering
\includegraphics[width=\linewidth]{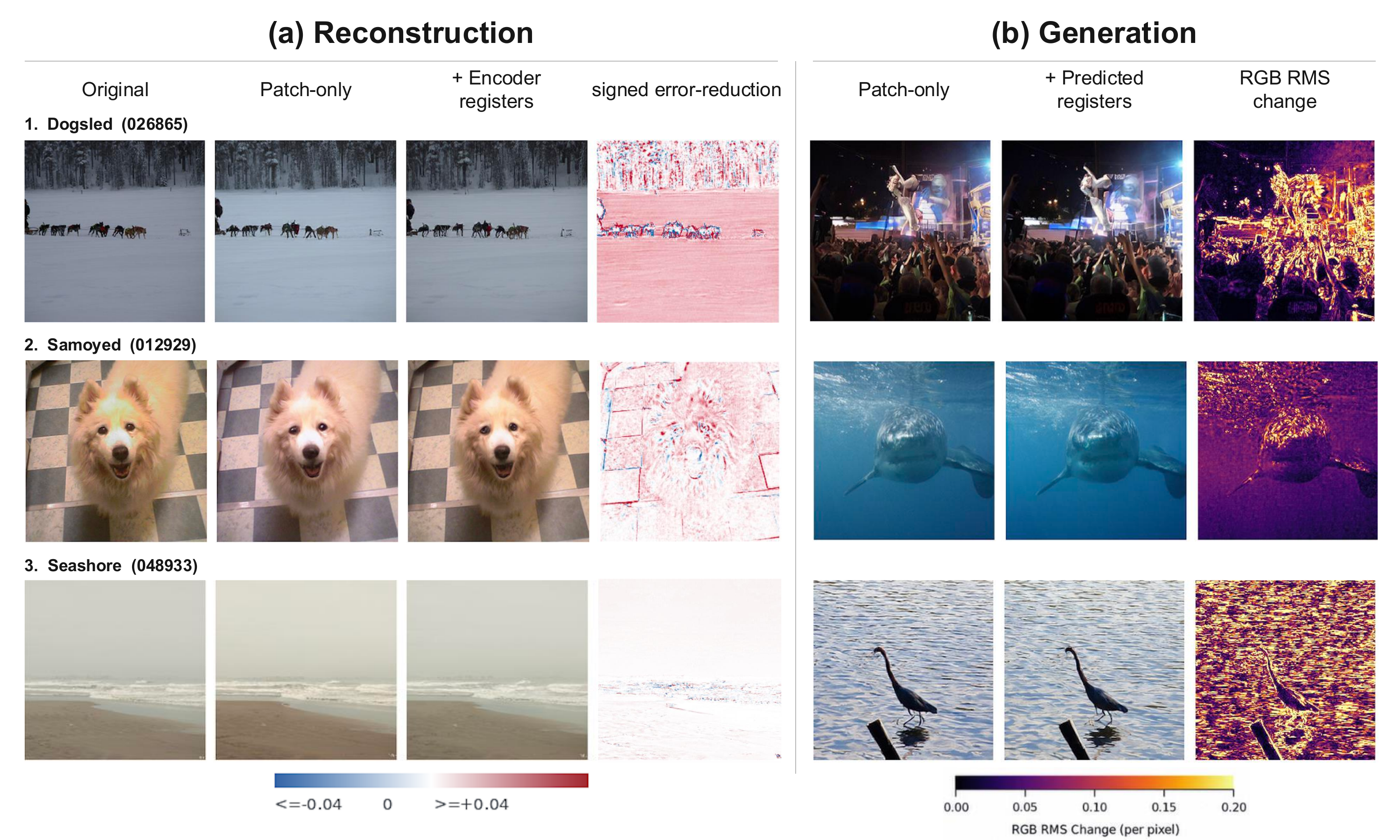}
\caption{\textbf{Reconstruction and generation with register inputs.}
(a) Original images, patch-only reconstructions, reconstructions
with native encoder registers, and signed reconstruction-error
reduction. Red indicates lower squared RGB error with registers;
blue indicates lower error with patches alone.
(b) Patch-only generation, generation with MHAP-predicted registers,
and the unsigned per-pixel RGB RMS difference between these outputs.
Each generation pair uses exactly the same stored patch tensor
from the pretrained DiT-DH/XL.
Generation difference maps show the magnitude of change, not
improvement, because no paired ground-truth image exists.
Reconstruction error maps use a fixed range of $[-0.04,0.04]$;
generation difference maps use $[0,0.20]$.
Examples are illustrative comparisons between separately trained
decoders. Model and sample identities appear in
Appendix~\ref{appendix:stage2-qualitative}.}
\label{fig:reconstruction-generation}
\end{figure}

\subsection{Register recovery and image generation}
\label{sec:stage2-generation}

\paragraph{Native versus predicted registers.}
Before evaluating generation, we test register recovery on
50,000 encoded ImageNet-1K validation images. We hold one
DINOv2 register-input decoder and each image's complete patch
sequence fixed, changing only the register source. DISTS is
0.1044 with native registers, 0.1121 with MHAP predictions,
0.1289 with the training-set mean, and 0.1333 with shuffled
predictions. MHAP therefore recovers image-specific features
useful for decoding, but native registers provide better
reconstruction fidelity. Most of the RGB MSE increase when
predictions replace native registers occurs in the channel-mean
term (Appendix~\ref{appendix:fixed-decoder-register-recovery}).

\paragraph{Generation from identical sampled patches.}

We decode the same 50,000 DINOv2 patch samples from a released
DiT-DH/XL using patch-only ViT-B, ViT-B with recovered registers,
and the released ViT-XL decoder (Table~\ref{tab:stage2-in1k}).
The generator, sampled tensors, and class labels are unchanged;
Figure~\ref{fig:reconstruction-generation}(b) shows paired decodings.

\begin{table}[t]
\centering
\small
\begin{tabular}{llc}
\toprule
Decoder & Decoding inputs & gFID $\downarrow$\\
\midrule
ViT-B & Patches & $1.8191\pm0.0058$\\
ViT-B & Patches + recovered registers
      & $\mathbf{1.6414}\pm0.0133$\\
\midrule
Released ViT-XL & Patches & 1.5572\\
\bottomrule
\end{tabular}
\caption{\textbf{Decoding identical pretrained-XL samples.}
ViT-B results are mean $\pm$ sample standard deviation across three
decoder seeds, with one fixed MHAP predictor for register recovery.
The released ViT-XL decoder is a larger-model reference.}
\label{tab:stage2-in1k}
\end{table}

On identical pretrained-DiT patch samples, decoding with
MHAP-predicted registers achieves mean gFID 1.6414,
compared with 1.8191 for the patch-only reference,
a reduction of 9.77\%.
All three register-decoder runs score below all three
patch-only runs.
The larger released decoder achieves gFID 1.5572.
This comparison uses unchanged diffusion samples, without
another denoising pass.
Generated images have no paired source, so PSNR and DISTS
are not used as generation-quality scores.

\section{Discussion}
\label{sec:discussion}

Our analysis shows that register tokens encode global scene information. We showed this quantitatively in Section~\ref{sec:probe-complementarity}, where registers improved the prediction of global image properties such as color histograms in comparison to using mean-pooled patch tokens alone. We also showed it qualitatively by training a decoder only on the register tokens. The reconstructions from registers alone recover aspects of global appearance, such as the color and texture of grass or clouds, while losing substantial object and spatial detail (see Figure~\ref{fig:register-only-original-comparison}). 

Current state-of-the-art decoders, such as those used in Representation Autoencoders (RAEs), take only the patch tokens of a pretrained encoder as input. We therefore tested whether the appearance information identified by our probing experiments also improves reconstruction quality when register tokens are given to the decoder as an additional input. Retaining four encoder registers alongside all spatial patches reduces mean ImageNet-1K reconstruction FID (rFID) by 9.5\% with DINOv2 and 5.2\% with DINOv3 relative to patch-only decoding, while also improving PSNR. The same decoders also improve out-of-distribution reconstruction without additional training: rFID decreases by 6.2\% and 5.2\% on NINCO and by 43.9\% and 28.1\% on Places365 for DINOv2 and DINOv3, respectively, alongside improved PSNR and DISTS.

During generation with RAEs, we recover the register tokens from the generated patch tokens using a trained MHAP predictor. This decouples the choice of decoder inputs from the choice of diffusion variables. For DINOv2, recovering registers from sampled patches reduces gFID by 9.8\% relative to patch-only decoding, without changing the pretrained diffusion model or its samples. Registers can therefore improve image decoding without becoming additional diffusion targets. Our ablations show that correspondence matters: within-class shuffling preserves
the register-bank distribution in each conditioning class but
worsens gFID with the decoder fixed (Appendix~\ref{appendix:generated-register-controls}). The internal-supervision comparisons also favor the MHAP route over the tested alternatives, including the larger InfoNCE decoder (Appendix~\ref{appendix:register-supervision}).

While prior work has indicated that register tokens encode global image properties~\citep{darcet2023registers, fel2025rabbithull}, our global-appearance probes and qualitative reconstruction analysis offer further insights into the global properties that they encode, and why they may fall short of the CLS and patch tokens when probed for semantic class information \citep{baxevanakis2026reassessregisters}. The improved classification performance reported by \citet{darcet2023registers} may therefore stem from this separation: with global background information absorbed by the register tokens that are not used in the probe, the CLS and patch tokens retain more disentangled semantic class information and are less affected by background cues, which can act as confounders during classification \citep{Lapuschkin2019}. Registers may thus be dispensable for classification, but not in general. We show this for image decoding and generation in RAEs, where retaining registers improves both. Beyond this, registers can also be used directly for applications such as global scene style transfer through spherical interpolation between register tokens (Figure~\ref{fig:register-appearance-control}).

Future work could explore additional applications in which access to global appearance and scene information is valuable, such as domain generalization~\citep{zhou2022domain}, where background or image style often acts as a confounder. Another direction is exploring alternative architectures and loss objectives that integrate register tokens more directly into the decoder. Finally, analyzing which information each register stores, and whether more registers lead to better disentanglement or greater information content, would further deepen our understanding of their role. 

\subsection*{AI use statement}
We used ChatGPT and Claude for assistance with conceptual framing, experimental
planning, literature discovery, drafting, and LaTeX preparation. 
Codex assisted with model and evaluation implementation,
debugging, configuration checks, and figure-export scripts. 
The authors take responsibility for the final manuscript,
code, figures, and scientific claims.

\bibliographystyle{iclr2027_conference}
\bibliography{references}
\clearpage
\appendix

\section{Experimental Details}
\label{appendix:exp-details}

\subsection{Feature coordinates and reconstruction objective}
\label{appendix:feature-loss-details}

The specifications below describe the 16-epoch ImageNet-1K
native-register decoders.

\paragraph{Frozen features and image preprocessing.}
The frozen checkpoint identifiers and revisions are:
\begin{itemize}
\item DINOv2: \texttt{facebook/dinov2-with-registers-base},\\
revision \texttt{a1d738ccfa7ae170945f210395d99dde8adb1805}.
\item DINOv3: \texttt{facebook/dinov3-vitb16-pretrain-lvd1689m},\\
revision \texttt{5931719e67bbdb9737e363e781fb0c67687896bc}.
\end{itemize}
Their actual encoder inputs are $224\times224$ and $256\times256$,
respectively, producing $16\times16$ patch grids with 768 channels
and four ordered register tokens. We extract the final
\texttt{last\_hidden\_state}, after the encoder's terminal LayerNorm.
This LayerNorm retains channel normalization but has its learned
scale and bias removed. Its variance is computed over the 768 channels
of each token, with epsilon $10^{-6}$ for DINOv2 and $10^{-5}$ for DINOv3.
The four registers retain their encoder order; the encoder CLS token
is excluded from the native-register decoder inputs.

For decoder training, RGB images are resized with PIL bicubic
interpolation to a shorter side of 384 pixels, rounding the other
side down, and randomly cropped to $256\times256$. The crop is
reproducible from the decoder seed, epoch, and image identity.
Evaluation and normalization-statistic fitting use a fixed center crop:
repeated BOX downsampling halves both dimensions while the shorter
side is at least 512, followed by bicubic resizing to a shorter side
of 256 (rounding each dimension to the nearest integer) and a centered
$256\times256$ crop. Pixels are divided by 255. Before encoding,
DINOv2 inputs are resized from 256 to 224 using tensor bicubic
interpolation with \texttt{align\_corners=False} and no antialiasing;
DINOv3 retains the $256\times256$ input. Both encoders use channel
means $(0.485,0.456,0.406)$ and standard deviations
$(0.229,0.224,0.225)$.

\paragraph{Coordinates and decoder positions.}
Let $P_{p,c}$ and $R_{j,c}$ denote terminal patch and register features.
The receiving RAE's stored patch-coordinate transform is
\begin{equation}
 S_P(P)_{p,c}=
 \frac{P_{p,c}-\mu^P_{p,c}}{\sqrt{v^P_{p,c}+10^{-5}}},
 \qquad
 S_R(R)_{j,c}=
 \frac{R_{j,c}-\mu^R_{j,c}}{\sqrt{v^R_{j,c}+10^{-5}}}.
\end{equation}
The means and population variances are fitted over all 1,279,690
training images, independently for every patch position and channel
or register slot and channel. Thus patch statistics have shape
$768\times16\times16$ and register statistics have shape $4\times768$.
For each coordinate $a$, $\mu_a=N^{-1}\sum_n f_{n,a}$ and
$v_a=N^{-1}\sum_n f_{n,a}^2-\mu_a^2$; accumulation uses FP64, stored
statistics use FP32, and negative roundoff in $v_a$ is clipped to zero.
These statistics remain fixed during evaluation.

Patch latents are stored or passed through $S_P$ and inverted before
the decoder's patch projection. Native reconstruction passes the raw
registers directly to the register projection, without $S_R$.
MHAP target standardization uses its separately fitted predictor
statistics, described with Equation~\ref{eq:register-prediction-loss}.
Its output is first converted back to encoder-feature coordinates.
The generation wrapper applies $S_R$ for the receiving RAE's interface,
and the decode call immediately applies $S_R^{-1}$; the transformer
therefore receives predicted registers in the same raw coordinates
as native registers, up to floating-point roundoff. Generated patch
tensors likewise undergo the source generator's inverse coordinate
transform before the receiver's patch-coordinate conversion.

The decoder sequence contains its own learned class token, followed
by four register tokens and then 256 patch tokens in raster order.
This decoder class token is distinct from the omitted encoder CLS.
Separate learned linear maps project patch and register features to
width 768. Patch positions use fixed two-dimensional sinusoidal
embeddings. If $e_j$ denotes the stored learned positional vector
for register slot $j$, its effective input is
$W_R R_j+b_R+2e_j$; $e_j$ is initialized from
$\mathcal N(0,0.02^2)$ and is added in both auxiliary-token preparation
and sequence-position addition. The same convention is used during
training and inference. Reconstruction pixels are predicted only
from the 256 patch positions.

\paragraph{Reconstruction objective.}
For a batch of RGB targets $x$ and reconstructions $\hat x$, the image
objective is
\begin{equation}
\mathcal L_{\mathrm{image}}
=
\mathcal L_{\mathrm{RGB}}
+
\mathcal L_{\mathrm{perc}}
+
0.05\,\mathcal L_{\mathrm{Gram}}
+
w_{\mathrm{adv}}(t)\mathcal L_{\mathrm{adv}}.
\label{eq:appendix-image-objective}
\end{equation}
Here $\mathcal L_{\mathrm{RGB}}=\operatorname{mean}|\hat x-x|$ averages
over the batch, channels, and pixels. The perceptual term is frozen
VGG16 LPIPS~\citep{johnson2016perceptual,zhang2018perceptual}, using
\texttt{relu1\_2}, \texttt{relu2\_2}, \texttt{relu3\_3},
\texttt{relu4\_3}, and \texttt{relu5\_3} (feature indices
$3,8,15,22,29$). For this term, the VGG input transform is
$((2x-1)-s)/q$, where $s=(-0.030,-0.088,-0.188)$ and
$q=(0.458,0.448,0.450)$. At every spatial position, each feature vector
$f$ is normalized as $f/(\sqrt{\sum_c f_c^2}+10^{-10})$.
The loss sums the five layers' squared feature differences weighted
by the frozen pretrained LPIPS $1\times1$ channel projections,
averages spatially within each layer, and averages over the batch.
LPIPS is in evaluation mode, so its dropout is inactive.

For the Gram term, let $F_\ell(x)\in\mathbb R^{C_\ell\times H_\ell W_\ell}$
be flattened, unnormalized VGG16 activations at layers
$\ell\in\{3,8,15\}$. Its VGG inputs are first clipped to $[0,1]$ and
normalized with the ImageNet channel means and standard deviations
above. Define
\begin{equation}
 G_\ell(x)=\frac{F_\ell(x)F_\ell(x)^\top}{C_\ell H_\ell W_\ell},
 \qquad
 \mathcal L_{\mathrm{Gram}}
 =\sum_{\ell\in\{3,8,15\}}
 \operatorname{mean}\left|G_\ell(\hat x)-G_\ell(x)\right|.
\end{equation}
Each mean is over batch members and all entries of the Gram matrix;
the three layer terms are summed with unit weights. The RGB and
LPIPS terms use the unclipped decoder output.

The decoder adversarial term is the negative mean discriminator
logit on augmented $2\hat x-1$. It activates after eight completed
epochs. Writing
$\mathcal L_0=\mathcal L_{\mathrm{RGB}}+\mathcal L_{\mathrm{perc}}
+0.05\mathcal L_{\mathrm{Gram}}$ and $W_{\mathrm{out}}$ for the final
pixel-prediction matrix, its weight is
\begin{equation}
 w_{\mathrm{adv}}(t)=0.75\,\mathbf 1[t\geq8]\,
 \operatorname{stopgrad}\!\left[
 \operatorname{clip}_{[0,1000]}
 \frac{\|\nabla_{W_{\mathrm{out}}}\mathcal L_0\|_2}
 {\|\nabla_{W_{\mathrm{out}}}\mathcal L_{\mathrm{adv}}\|_2+10^{-6}}
 \right].
\end{equation}
There is no register-prediction term in the native-register decoder
objective. MHAP is trained separately.

\paragraph{Optimizer and schedule.}
These decoders use AdamW with betas $(0.9,0.95)$, weight decay zero,
global batch size 256, BF16 computation, and no gradient clipping.
Let $u$ be the number of completed optimizer steps and
$t=u/4998$ the epoch-equivalent training time. The learning rate,
applied before each step, is
\begin{equation}
 \eta(t)=\begin{cases}
 2\times10^{-4}\,t, &0\leq t<1,\\
 2\times10^{-5}+\frac{2\times10^{-4}-2\times10^{-5}}{2}
 \left[1+\cos\!\left(\pi\frac{t-1}{15}\right)\right],
 &1\leq t\leq16.
 \end{cases}
\end{equation}
Training uses 4,998 complete batches per epoch and 79,968 steps in
total. EMA decay is 0.9978, and evaluation uses the final 16-epoch
EMA export. Clean runs use zero patch noise, zero register noise,
and zero register dropout. Evaluation outputs are clipped to
$[0,1]$, multiplied by 255, rounded, and stored as RGB uint8 images.

\subsection{Evaluation and uncertainty}
\label{appendix:evaluation-uncertainty}

The ImageNet-100 appearance experiment evaluates 5,000 validation
images for each decoder seed. ImageNet-1K reconstruction evaluates
50,000 validation images. Paired comparisons use the same image
identities, preprocessing, and output conversion for both models.

For an image-level metric, we first average over the evaluation
images within each decoder seed. We then compute the mean and sample
standard deviation across decoder seeds. Paired confidence intervals
are computed from the within-seed differences between models.
With three seeds, the 95\% interval is
\begin{equation}
\overline{\Delta}
\;\pm\;
t_{0.975,2}\frac{s_{\Delta}}{\sqrt{3}},
\label{eq:appendix-seed-ci}
\end{equation}
where $s_{\Delta}$ is the sample standard deviation of the three
seed-level differences. These intervals describe training-seed
variation; evaluation images are not treated as independent
training replicates. Tables retaining a separately reported
standard-deviation convention identify it in their captions.

FID compares image distributions in Inception feature
space~\citep{heusel2017gans}. PSNR measures paired pixel fidelity,
and DISTS measures paired perceptual differences in structure and
texture~\citep{ding2021dists}. FWD compares distributions of
wavelet-packet features
(Appendix~\ref{appendix:appearance-metric-sensitivity}).

Supplementary FD-DINOv2 measurements use DINOv2 features and are
denoted rFDD for reconstruction and gFDD for generation. They are
raw distributional distances, distinct from paired feature errors.
References and feature-extraction procedures are held fixed within
each reported metric comparison.

\paragraph{PSNR aggregation.}
PSNR is computed separately for each reconstructed image and then
averaged over the evaluation set.
For RGB values in $[0,1]$,
\[
\overline{\operatorname{PSNR}}
=
-\frac{10}{n}\sum_{i=1}^{n}\log_{10}\operatorname{MSE}_i.
\]
This differs from applying the logarithm to arithmetic-mean MSE.
We therefore calculate relative reductions in mean pixel error
directly from the MSE measurements rather than converting an
average PSNR difference into an arithmetic-mean error reduction.
The channel-mean and centered contributions are additive in MSE,
not in PSNR. DISTS differences are reported in the metric's own
units, rather than interpreted as percentages of perceived quality.

\subsection{Compute and inference accounting}
\label{appendix:compute}

Register recovery evaluates MHAP once per sampled patch representation,
followed by the image decoder; it adds no diffusion sampling pass.
MHAP contains 15,366,144 parameters. We measure the patch-only and
MHAP-plus-register routes on one NVIDIA A100-PCIE-40GB with BF16
autocast, using the epoch-16 decoders and the MHAP predictor.

\begin{table}[t]
\centering
\small
\begin{tabular}{rlrr}
\toprule
Batch & Route & Latency (ms/batch) & Peak memory (MiB)\\
\midrule
1 & Patch-only & 9.610 & 377.3\\
1 & MHAP + registers & 11.577 & 439.3\\
8 & Patch-only & 9.420 & 477.5\\
8 & MHAP + registers & 11.717 & 541.4\\
\bottomrule
\end{tabular}
\caption{\textbf{Decoder-stage inference cost.}
Median CUDA-event latency per batch from 100 forwards after 20 warmups.
Each route and batch size uses a fresh process; the two routes use
identical stored GPU-resident patch inputs at each batch size.
Peak allocated memory includes route weights, inputs, temporaries,
and the live output, but excludes allocations outside PyTorch's
CUDA allocator.}
\label{tab:inference-cost}
\end{table}

At batch sizes 1 and 8, MHAP-plus-register decoding increases latency
by 20.47\% and 24.38\%, and peak allocated memory by 16.42\% and 13.38\%,
respectively. Measurements include input-coordinate conversions and GPU
uint8 image conversion. They exclude diffusion sampling, image encoding,
model loading, and external input/output transfers. These measurements
characterize one model pair; repeated forwards do not estimate variation
across trained models.

\section{Additional Appearance Probes}
\label{appendix:full-probes}

\subsection{Individual-register and register-bank probes}

Table~\ref{tab:dino-conditional-full} compares mean-pooled patches,
patches with one register, and patches with the complete register bank.
The register identified as \texttt{reg4} was selected from the
appearance-probe analysis rather than on a separate selection set.
Its result therefore describes the selected register, not an
unbiased estimate of selecting a register for a new dataset.

The complete bank gives the highest thumbnail and nonzero
low-frequency DCT scores. The selected individual register gives
higher color-histogram and DC scores than the complete bank.
Adding more register features does not therefore improve every
linear prediction target.

\begin{table}[t]
\centering
\small
\begin{tabular}{lrrr}
\toprule
Target & Patch mean & +\texttt{reg4} & +Register bank\\
\midrule
Thumbnail       & 0.211 & 0.223 & 0.228\\
Color histogram & 0.288 & 0.316 & 0.296\\
DCT DC          & 0.735 & 0.827 & 0.820\\
DCT AC          & 0.038 & 0.040 & 0.068\\
\bottomrule
\end{tabular}
\caption{\textbf{Individual-register and register-bank probes.}
Held-out $R^2$ from mean-pooled patches, with the selected register,
or with all four registers. DC denotes the spatial mean; AC denotes
nonzero low-frequency DCT coefficients.}
\label{tab:dino-conditional-full}
\end{table}

\subsection{Full-patch appearance probes}
\label{appendix:full-patch-probe}

A separate experiment fits regularized linear probes to the flattened
full patch sequence, using online BF16 feature extraction and nested
cross-validation. Unlike mean pooling, flattening preserves the
positions of all patch features in the probe input.
Table~\ref{tab:full-patch-appearance} reports all evaluated frequency
bands and the thumbnail target. Absolute scores are not directly
comparable with the patch-mean experiment because the extraction
and fitting protocols differ.

\begin{table}[t]
\centering
\small
\begin{tabular}{lrrr}
\toprule
Target & Full patches & +Registers & $\Delta R^2$\\
\midrule
RGB DC / global mean & 0.57336 & 0.66280 & +0.08944\\
Very-low DCT         & 0.46391 & 0.46386 & -0.00005\\
Low DCT              & 0.16085 & 0.16082 & -0.00003\\
Mid DCT              & 0.00466 & 0.00437 & -0.00029\\
High DCT             & -0.00237 & -0.00242 & -0.00005\\
Thumbnail            & 0.24129 & 0.24711 & +0.00582\\
\bottomrule
\end{tabular}
\caption{\textbf{Full-patch appearance probes.}
Held-out $R^2$, averaged across three shared outer splits.
Changes are computed before rounding.}
\label{tab:full-patch-appearance}
\end{table}

The largest improvement is in mean RGB prediction:
DC $R^2$ increases by 0.08944, while thumbnail $R^2$
increases by 0.00582. Aggregate non-DC $R^2$ changes
by $-0.00019$.

With identical splits and effective regularization candidates,
zero-register inputs reproduce the patch-only predictions
exactly. Shuffled-register and Gaussian-register ablation experiments give
DC increments of approximately zero and $-0.00095$,
respectively; their thumbnail increments are $-0.00083$
and $-0.00038$.

The matched experiment uses 5,000 images and three shared
outer splits, with identical inner splits across conditions.
Separately row-normalized patch and register features define
the additive kernel $K_P+\lambda K_R$, keeping patch-kernel
scaling fixed. Regularization and auxiliary-kernel weight are
selected within the training data for each target coordinate.
Encoder revision, preprocessing, image identities, features,
and split assignments are retained.

\subsection{Reconstruction from registers alone}
\label{appendix:register-only-reconstruction}

\begin{figure}[!htbp]

\centering
\includegraphics[
  width=\linewidth,
  height=0.72\textheight,
  keepaspectratio
]{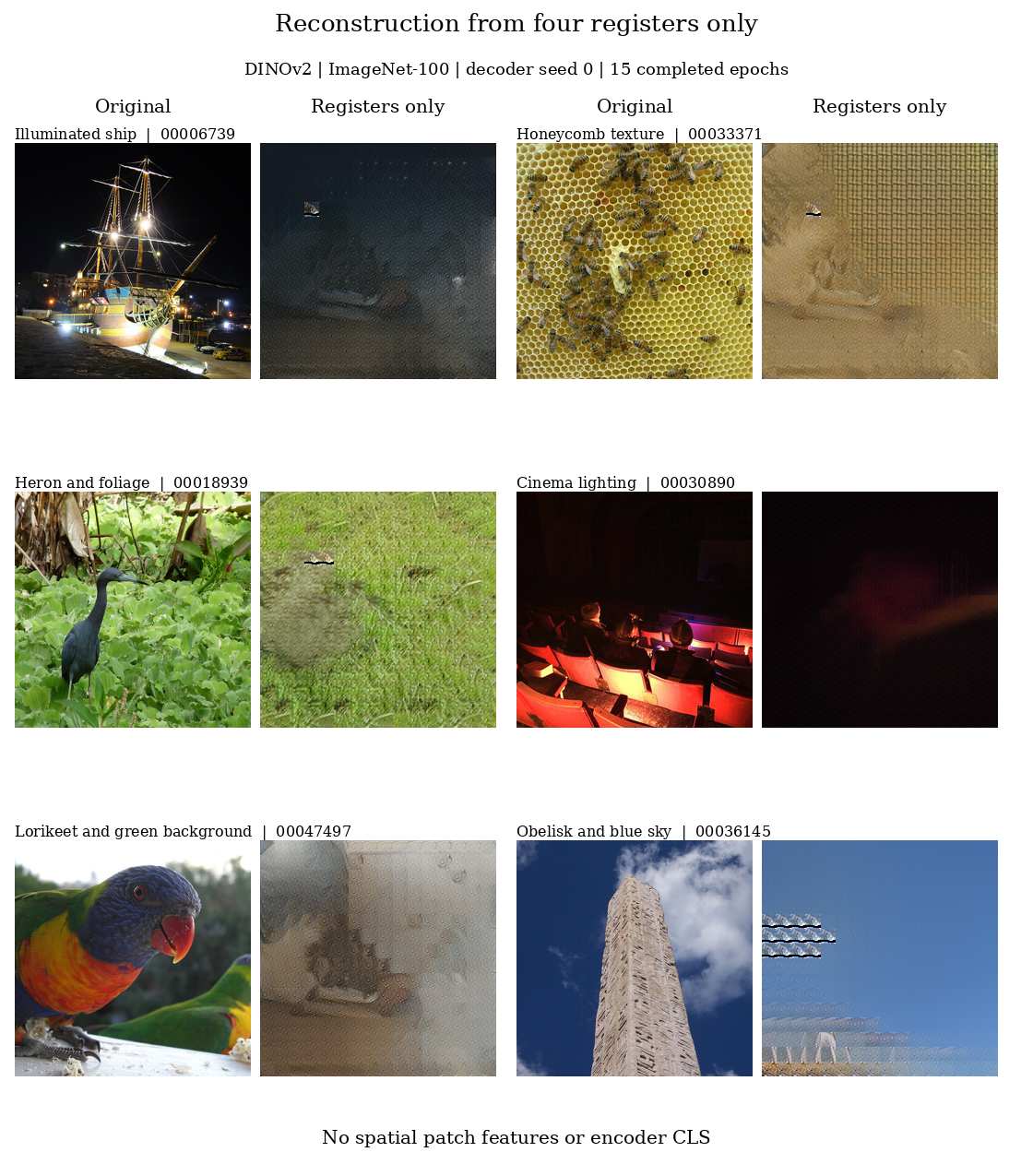}
\caption{\textbf{Additional reconstructions from four registers alone.}
Each adjacent pair shows an original photograph and its register-only
reconstruction. Examples show an illuminated ship, honeycomb texture,
foliage, cinema lighting, plumage, and blue sky. The ImageNet-100-trained DINOv2
ViT-B decoder receives four encoder registers, without spatial patch
features or encoder CLS. All outputs use decoder seed 0 and EMA
weights after 15 completed epochs.}
\label{fig:appendix-register-only}
\end{figure}

The separate DINOv2 ViT-B decoder in
Figure~\ref{fig:register-only-original-comparison} receives the
encoder's four registers, without spatial patch features
or encoder CLS. The four register positions are interpolated to
256 decoder positions to reconstruct a $256\times256$ image.
The decoder is trained on ImageNet-100 with seed 0; the displayed
outputs use EMA weights after 15 completed epochs of a 16-epoch
schedule. The main-paper curtain, aircraft-wing, terrier, window-screen,
and overcast waterfront are selected after inspecting the outputs. 
Their full image extent and decoded colors are retained.
Figure~\ref{fig:appendix-register-only} adds six examples from the
same register-only decoder.

\section{Decoder Ablations and Reference Results}
\label{appendix:stage1-controls}

\subsection{Alternative register integration}
\label{appendix:integration}

The principal decoder prepends register features to the patch
sequence and processes both through joint self-attention.
Two alternative interfaces expose the same type of non-spatial
input through modulation or cross-attention.

Pooled modulation compresses the register bank into a vector
$q=\operatorname{Pool}(Z_r)$ and predicts a feature-wise affine
transformation:
\begin{equation}
(\gamma,\beta)=g(q),
\qquad
\widetilde h_\ell=\gamma\odot h_\ell+\beta.
\label{eq:appendix-register-modulation}
\end{equation}
Cross-attention instead uses projected register features as a
memory, with queries from the patch stream.

These interfaces differ in how register information enters the
decoder. Modulation conditions patch features through a pooled
vector; cross-attention reads a separate register memory; prepending
allows patches and registers to evolve jointly. Results involving
CLS-plus-register inputs are identified separately from the
four-register-only comparison.

\subsection{Choice of non-spatial decoder inputs}
\label{appendix:input-choice}

We compare decoding from patches alone with the addition of
CLS, four registers, or their combination.
All conditions retain the complete spatial patch sequence.
Table~\ref{tab:nonspatial-input-choice} reports the reconstruction
FID for each encoder's input study.

\begin{table}[t]
\centering
\small
\begin{tabular}{lrr}
\toprule
Decoder input & DINOv2 rFID $\downarrow$
              & DINOv3 rFID $\downarrow$\\
\midrule
P & 4.8388 & 4.49223\\
P + CLS & 4.7474 & 4.37634\\
P + REG & 4.6113 & 4.33774\\
P + REG + CLS & 4.5932 & 4.33395\\
\bottomrule
\end{tabular}
\caption{\textbf{Non-spatial decoder inputs.}
Reconstruction FID in the DINOv2 and DINOv3 input studies.
P denotes the complete spatial patch sequence and REG denotes
the four encoder registers. Comparisons are made within each
encoder's experiment.}
\label{tab:nonspatial-input-choice}
\end{table}

Register inputs yield lower rFID than CLS inputs in both studies.
Adding CLS to the register bank reduces rFID by a further 0.0181
for DINOv2 and 0.00379 for DINOv3, corresponding to 0.39\% and
0.09\% of the respective register-input scores.
This pattern is consistent with the appearance-related specialization
of register features described by
\citet{fel2025rabbithull} and motivates the four-register interface
used in our principal experiments.
The decoder-capacity study below evaluates the combined
CLS-plus-register input and retains that input definition.

\subsection{Decoder capacity}
\label{appendix:decoder-capacity}

The capacity experiment asks whether a larger patch-only decoder
removes the advantage of supplying non-spatial encoder outputs.
The ImageNet-100 comparison evaluates ViT-B, ViT-L, and
ViT-XL decoders with DINOv2 features. The augmented models receive
the encoder CLS token together with four registers, in addition to
all spatial patches. This is a five-token auxiliary-input comparison,
not a four-register-only ablation.

\begin{table}[t]
\centering
\small
\begin{tabular}{lrrrr}
\toprule
Decoder & Epochs & Patches & +CLS+4REG & Reduction\\
\midrule
ViT-B  & 16 & 5.7289 & 5.3722 & 6.2\%\\
ViT-L  & 16 & 5.4250 & 5.1241 & 5.5\%\\
ViT-XL & 16 & 5.4720 & 5.1724 & 5.5\%\\
\midrule
ViT-L  & 40 & 4.6091 & 4.4234 & 4.0\%\\
\bottomrule
\end{tabular}
\caption{\textbf{Decoder capacity with DINOv2 features on ImageNet-100.}
Reconstruction FID for patches alone and for patches with CLS and
four registers. Epochs denote the recorded training-schedule length.
The 16-epoch B/L/XL values are recorded manuscript measurements
without retained seed-level metric records. The 40-epoch ViT-L
row is a separately retained seed-0 evaluation.
Relative reductions are computed within each row.}
\label{tab:decoder-capacity}
\end{table}

The auxiliary-input decoder has lower rFID at each tested size.
Increasing capacity from ViT-B to ViT-L improves the patch-only
reference, but the auxiliary-input advantage remains.
ViT-XL does not further improve the recorded ViT-L scores.
The advantage persists across the tested capacity range:
6.2\%, 5.5\%, and 5.5\% for ViT-B, ViT-L, and ViT-XL,
respectively.

The longer ViT-L comparison also improves paired fidelity:
PSNR increases from 18.9101 to 19.2504\,dB and SSIM from
0.4706 to 0.4790. Its clean-cache rFDD decreases from 37.4515
to 35.9324. These measurements use the combined CLS-plus-register input.
Four-register-only capacity effects and training-seed uncertainty
are unavailable for the B/L/XL sweep.

\subsection{Register supervision versus register inputs}
\label{appendix:register-supervision}

We compare two ways of using four terminal encoder registers
as training targets with supplying registers through the decoder
interface. All rows use DINOv2 ViT-B features and decoder seed 0
at the 16-epoch ImageNet-1K endpoint, with $\tau=0$.
Reconstruction uses the same 50,000 validation images.
Generation uses the same retained 50,000 pretrained-XL patch
samples and evaluation reference.

\paragraph{Internal register supervision.}
The contrastive variant uses four learned slots with bidirectional
cross-attention between patches and slots. After two completed
decoder blocks, a LayerNorm--linear head predicts the four
standardized encoder registers. Its additional objective is
\[
0.05\,\mathcal{L}_{\mathrm{InfoNCE}}
+0.05\,\mathcal{L}_{\mathrm{slot}}.
\]
InfoNCE compares corresponding register slots across images,
with temperature $0.07$.
The slot regularizer combines a variance hinge with target
standard deviation $0.5$ and the mean squared cosine similarity
between distinct slots. Slot dropout is $0.1$ during training.
The prediction head is removed for inference; the learned slots
and cross-attention remain. Patches are the only image-dependent
inference input.

The register-MSE variant instead prepends four learned slots.
After two completed self-attention blocks, a LayerNorm--linear
head predicts four standardized registers. A learned linear
projection maps these predictions back to decoder width and
replaces the four slot states before the remaining ten blocks.
The register MSE has coefficient $0.1$.
The prediction head and projection remain in the image path,
so the image objective also trains this prediction pathway.
This variant uses neither native registers nor an external
predictor at inference, and uses no slot dropout.

\begin{table}[t]
\centering
\small
\setlength{\tabcolsep}{4pt}
\begin{tabular}{lrrr}
\toprule
Approach & Parameters (r/g)
& rFID $\downarrow$ & gFID $\downarrow$\\
\midrule
Patch-only
& 86.435 / 86.435 & 0.5253 & 1.8189\\
Internal InfoNCE
& 143.806 / 143.806 & 0.5104 & 1.7888\\
Internal register MSE
& 88.215 / 88.215 & 0.5181 & 1.8399\\
Native / MHAP registers
& 87.029 / 102.395 & \textbf{0.4708} & \textbf{1.6561}\\
\bottomrule
\end{tabular}
\caption{\textbf{Register supervision and register inputs.}
DINOv2 ImageNet-1K, decoder seed 0, 16 epochs.
Parameter counts are in millions for reconstruction/generation,
excluding the frozen encoder, diffusion generator, and
training-only prediction head. Counts include fixed positional
parameters and, for InfoNCE, 3,072 stored null-slot parameters
unused during ordinary inference.}
\label{tab:register-supervision}
\end{table}

The contrastive variant improves both recorded FIDs relative
to the patch-only reference. The register-MSE variant improves
reconstruction FID but increases generation FID.
Neither reaches the scores obtained with native registers for
reconstruction and MHAP predictions for generation.
These comparisons show that the tested uses of register
supervision and register inputs produce different outcomes.
Because decoder architectures and objectives also differ,
they do not isolate the effect of loss choice or establish
a parameter-matched comparison.

\subsection{Learned decoder context and patch summaries}
\label{appendix:summary-controls}

The decoder-context ablations add five tokens without supplying
encoder CLS or register outputs. One condition uses learned
image-independent tokens. The other produces five
\textsc{PatchMAP} summary tokens from the spatial patch sequence.
Table~\ref{tab:dino-summary-controls} compares both with their
own seed-0 patch-only reference.

\begin{table}[t]
\centering
\small
\begin{tabular}{lcc}
\toprule
Decoder input & rFID $\downarrow$ & rFDD $\downarrow$\\
\midrule
Patches & 4.8284 & 41.3454\\
Patches + 5 learned tokens & 4.7992 & 41.3672\\
Patches + 5 \textsc{PatchMAP} summaries & 4.8168 & 40.9061\\
\bottomrule
\end{tabular}
\caption{\textbf{Decoder-context ablations on ImageNet-100.}
Single-seed measurements with five added tokens.
Neither context condition receives non-spatial encoder outputs.
The ablation decoders are not token-count matched to the four-register
decoder.}
\label{tab:dino-summary-controls}
\end{table}

Learned tokens reduce rFID by 0.0292 while increasing rFDD by
0.0218. Patch summaries reduce rFID by 0.0116 and rFDD by 0.4393.
These ablations use one seed and five tokens; the principal
register result uses four tokens and three decoder seeds.

\section{Supplementary Reconstruction Measurements}
\label{appendix:supplementary-reconstruction-measurements}

\subsection{ImageNet-100 decoder-input comparison}

The decoder-input study compares patches alone with patches and four
registers across three decoder-training seeds. All spatial patches are
retained and both conditions use the same base image objective.
These models are separate from the A/B appearance-analysis models in
Section~\ref{sec:appearance-effects}.

\begin{table}[t]
\centering
\small
\begin{tabular}{lc}
\toprule
Decoder input & rFID $\downarrow$\\
\midrule
Patches & $4.8388\pm0.0302$\\
Patches + registers & $\mathbf{4.6113}\pm0.0087$\\
\bottomrule
\end{tabular}
\caption{\textbf{ImageNet-100 decoder-input comparison.}
Mean and reported standard deviation across three decoder seeds.
Both routes receive all spatial patch features; the second also
receives the encoder's four register outputs.}
\label{tab:register-reconstruction}
\end{table}

\subsection{ImageNet-1K reconstruction by decoder seed}
\label{appendix:principal-reconstruction-seeds}

Table~\ref{tab:in1k-reconstruction-per-seed} supplies the per-seed
measurements for Table~\ref{tab:in1k-stage1}. Each reconstruction set
contains the same 50,000 validation image identities within an encoder
comparison. Means and sample standard deviations are computed across
the three decoder seeds. The aggregate rows reproduce the principal reconstruction
comparison in Table~\ref{tab:in1k-stage1}.

\begin{table}[t]
\centering
\small
\begin{tabular}{llcc}
\toprule
Encoder & Decoder seed
        & Patches rFID $\downarrow$
        & +Registers rFID $\downarrow$\\
\midrule
DINOv2 & 0 & 0.525288 & 0.470804\\
       & 1 & 0.516679 & 0.469105\\
       & 2 & 0.521390 & 0.475329\\
       & Mean $\pm$ SD
       & $0.521119\pm0.004311$
       & $0.471746\pm0.003217$\\
\midrule
DINOv3 & 0 & 0.464152 & 0.443066\\
       & 1 & 0.467069 & 0.440756\\
       & 2 & 0.469339 & 0.443407\\
       & Mean $\pm$ SD
       & $0.466853\pm0.002600$
       & $0.442410\pm0.001442$\\
\bottomrule
\end{tabular}
\caption{\textbf{ImageNet-1K reconstruction by decoder seed.}
All decoders use the ViT-B architecture. Register-input decoding
supplies four encoder registers and the complete patch sequence.
The aggregate rows reproduce the principal reconstruction comparison.}
\label{tab:in1k-reconstruction-per-seed}
\end{table}

\subsection{ImageNet-1K register-decoder fidelity}

Table~\ref{tab:in1k-register-fidelity} reports paired image fidelity
and wavelet distance for the register-input decoders.

\begin{table}[t]
\centering
\small
\begin{tabular}{lccc}
\toprule
Encoder & PSNR (dB) $\uparrow$ & DISTS $\downarrow$
        & FWD $\downarrow$\\
\midrule
DINOv2
& $19.7262\pm0.0012$
& $0.10444\pm0.00005$
& $0.4579\pm0.0105$\\
DINOv3
& $20.5406\pm0.0156$
& $0.09670\pm0.00007$
& $0.4443\pm0.0050$\\
\bottomrule
\end{tabular}
\caption{\textbf{Fidelity of ImageNet-1K register-input reconstructions.}
Mean $\pm$ sample standard deviation across three decoder seeds.
PSNR and DISTS compare each reconstruction with its source image;
FWD compares image distributions in wavelet coordinates.
Both rows use patches and four encoder registers.}
\label{tab:in1k-register-fidelity}
\end{table}

DINOv3's higher PSNR accompanies lower mean RGB MSE, 0.01134
versus 0.01334 for DINOv2. Its lower DISTS indicates closer agreement
with the source images under that structure-and-texture metric.

\section{Out-of-Distribution Reconstruction}
\label{appendix:ood-reconstruction}

\subsection{ImageNet-1K-trained decoders}
\label{sec:ood-reconstruction}

We evaluate ImageNet-1K-trained DINOv2 and DINOv3 decoders
on NINCO and Places365 without additional
training~\citep{bitterwolf2023inorout,zhou2018places}.
The evaluation sets contain 5,878 NINCO images and 36,500
Places365 images. Within each encoder comparison, both routes
use a ViT-B decoder backbone and the 16-epoch endpoint.
We compare decoding from the complete patch sequence with
decoding from patches and four native encoder registers.
Evaluation images, preprocessing, output conversion, and
metric implementations are held fixed across the two routes.
Table~\ref{tab:ood-register-reconstruction} reports
reconstruction FID and paired image fidelity.

\begin{table}[t]
\centering
\small
\setlength{\tabcolsep}{3pt}
\begin{tabular}{llccc}
\toprule
Encoder & Decoder input & rFID $\downarrow$
        & PSNR (dB) $\uparrow$ & DISTS $\downarrow$\\
\midrule
\multicolumn{5}{l}{\textit{NINCO}}\\
DINOv2
& Patches
& $4.9567$
& $19.9535$
& $0.12304$\\
& Patches + registers
& $4.6502$
& $20.2624$
& $0.11652$\\
DINOv3
& Patches
& $4.6931$
& $20.7924$
& $0.11497$\\
& Patches + registers
& $4.4501$
& $21.1945$
& $0.10909$\\
\midrule
\multicolumn{5}{l}{\textit{Places365}}\\
DINOv2
& Patches
& $2.5935$
& $19.1880$
& $0.11521$\\
& Patches + registers
& $1.4542$
& $19.3566$
& $0.10995$\\
DINOv3
& Patches
& $2.5663$
& $20.0437$
& $0.10823$\\
& Patches + registers
& $1.8441$
& $20.2301$
& $0.10334$\\
\bottomrule
\end{tabular}
\caption{\textbf{Reconstruction outside ImageNet-1K.}
ImageNet-1K-trained ViT-B decoders evaluated after 16 epochs,
without training on either evaluation dataset.
All results use decoder seed 0.}
\label{tab:ood-register-reconstruction}
\end{table}

Register-input decoding improves all three metrics for both
encoder families on both datasets.
For DINOv2, rFID decreases by 6.18\% on NINCO and
43.93\% on Places365.
The corresponding reductions for DINOv3 are 5.18\%
and 28.14\%.
All percentages are computed from unrounded scores.

Paired reconstruction fidelity improves alongside rFID.
On NINCO and Places365, respectively, DINOv2 gains
0.3089 and 0.1686\,dB in PSNR, while DISTS decreases
by 0.00652 and 0.00526.
DINOv3 gains 0.4022 and 0.1863\,dB in PSNR, with
DISTS reductions of 0.00588 and 0.00489.
These results show that the reconstruction benefits extend
to images outside the decoder-training dataset.

Among register-input decoders, DINOv3 achieves higher PSNR
and lower DISTS on both datasets.
The rFID ordering depends on the dataset: DINOv3 scores
lower on NINCO, whereas DINOv2 scores lower on Places365.

\subsection{ImageNet-100-trained patch-only decoders}

Table~\ref{tab:dino-ood-main} reports patch-only references for decoders
trained on ImageNet-100. These differ in training dataset from the
ImageNet-1K register decoders above and are not paired counterparts
to Table~\ref{tab:ood-register-reconstruction}.

\begin{table}[t]
\centering
\small
\begin{tabular}{lrr}
\toprule
Evaluation dataset & rFID $\downarrow$ & rFDD $\downarrow$\\
\midrule
ImageNet-R & 2.44 & 22.07\\
NINCO      & 6.62 & 65.76\\
Places365  & 4.88 & 26.30\\
\bottomrule
\end{tabular}
\caption{\textbf{ImageNet-100-trained patch-only OOD references.}
DINOv2 patch-only decoders trained on ImageNet-100.
The table preserves the reported reference scores without adding
unmeasured register-decoding comparisons.}
\label{tab:dino-ood-main}
\end{table}

\section{Register Recovery and Generation Details}
\label{appendix:recovery-generation}

\subsection{Image-space evaluation of recovered registers}
\label{appendix:fixed-decoder-register-recovery}
Before applying register recovery to generated patches, we
evaluate its image-space consequences on 50,000 ImageNet-1K
validation images. We hold one DINOv2 register-input decoder
and each image's complete patch sequence fixed, and compare
native encoder registers, MHAP predictions, training-set mean
registers, and shuffled MHAP predictions
(Table~\ref{tab:fixed-decoder-register-recovery}).
\begin{table}[t]
\centering
\small
\begin{tabular}{lccc}
\toprule
Register source
& \shortstack{Mean-color MSE\\($\times10^{-3}$)}
& \shortstack{Centered MSE\\($\times10^{-3}$)}
& DISTS $\downarrow$\\
\midrule
Encoder registers & 0.5665 & 12.7735 & 0.10441\\
MHAP predictions & 1.6368 & 13.0146 & 0.11205\\
Training-set mean & 4.5069 & 13.5449 & 0.12890\\
Shuffled MHAP & 6.2529 & 13.8718 & 0.13331\\
\bottomrule
\end{tabular}
\caption{\textbf{Register recovery with a fixed decoder.}
One decoder--predictor pair, 50,000 ImageNet-1K images, and identical
receiver patches. The MSE columns correspond to the two terms in
Equation~\ref{eq:appearance-centered-mse}; lower values indicate
smaller errors.}
\label{tab:fixed-decoder-register-recovery}
\end{table}

Predictions give lower pixel and perceptual errors than mean or
shuffled replacements, showing the value of receiver-specific recovery.
Encoder registers remain better: substituting predictions increases
DISTS by 0.00764 (paired image-bootstrap 95\% interval
$[0.00756,0.00772]$, conditional on the fixed models).
Mean-color error accounts for most of the total MSE increase.
High feature agreement therefore does not imply equivalent decoded
images (Appendix~\ref{appendix:posthoc-register-prediction}).

\subsection{Register replacements on generated patches}
\label{appendix:generated-register-controls}

We test register-input specificity using the existing 50,000 pretrained
DiT-DH/XL patch samples. The DINOv2 four-register decoder, MHAP predictor, sampled patches,
token positions, output processing, and official ADM ImageNet-256
population reference remain fixed. We compare aligned MHAP predictions
with the per-slot training-set mean and with within-class shuffled
predictions. Shuffling permutes complete four-register banks while
preserving slot order; permutation seed 0 produces zero self-assignments
and zero cross-class assignments. MHAP is evaluated before replacement
in both control routes.

\begin{table}[t]
\centering
\small
\begin{tabular}{lrr}
\toprule
Register source & gFID $\downarrow$ & Change from aligned\\
\midrule
Aligned MHAP & \textbf{1.6561} & ---\\
Within-class shuffled MHAP & 1.7981 & $+8.57\%$\\
Training-set mean & 2.3155 & $+39.81\%$\\
\bottomrule
\end{tabular}
\caption{\textbf{Register replacements during generation.}
One fixed decoder--predictor pair and identical sampled patches.
Relative changes use the aligned seed-0 score as denominator.}
\label{tab:generated-register-controls}
\end{table}

Aligned predictions yield lower FID than either replacement, supporting
the usefulness of sample-specific register values in the trained decoder.

\subsection{MHAP predictor}
\label{appendix:posthoc-register-prediction}

\paragraph{Architecture and inputs.}
We fit a separate predictor for each encoder, using its complete sequence
of 256 terminal patch features of dimension 768 to predict four register
features in encoder-defined order. A learned $768\!\times\!768$ projection
and LayerNorm transform the standardized patch features. Four learned
768-dimensional queries pass through two attention-pooling blocks, each
with 12-head cross-attention over the same patch sequence and a
$768\!\to\!3072\!\to\!768$ GELU feed-forward network. Attention and
feed-forward updates have residual connections and pre-LayerNorm;
queries and source features have separate attention-input LayerNorms.
A final LayerNorm and linear projection produce four 768-dimensional
outputs. No additional positional embeddings, query self-attention,
dropout, or feature noise are used. The predictor has 15,366,144
parameters and does not receive or predict CLS.

The frozen teacher uses terminal features with the final LayerNorm
affine parameters removed. Each source photograph receives the fixed
256-pixel center crop; DINOv2 inputs are then resized to $224\!\times\!224$
by bicubic interpolation, whereas DINOv3 uses $256\!\times\!256$.
Both use the pinned encoder processor's RGB mean and standard deviation.
Teacher features are extracted online in microbatches of 16 and retained
at a BF16 feature boundary. Predictor training is independent of the
image decoder and diffusion model.

\paragraph{Standardization and objective.}
Statistics are fitted separately for each encoder using only the
1,279,690-image scientific training subset. Patch standardization uses
one mean and standard deviation per feature channel, pooled across
images and all 256 patch positions. In particular, if $\mu_{cs}$ and
$v_{cs}$ are the population moments at channel $c$ and position $s$,
\begin{equation*}
\bar\mu_c=\frac{1}{256}\sum_s\mu_{cs},\qquad
\bar v_c=\frac{1}{256}\sum_s(v_{cs}+\mu_{cs}^2)-\bar\mu_c^2.
\end{equation*}
This includes between-position variation. Register targets use separate
population moments for each of the $4\!\times\!768$ slot--channel
coordinates, pooled over images. In both cases the divisor is
$\max(\sqrt{\max(v,0)},10^{-6})$: the predictor floors the standard
deviation rather than adding an epsilon inside the square root.
Population moments use FP64 accumulation and are saved in FP32.

Let $\mathcal S_p$ and $\mathcal S_r$ denote these patch and register
standardizations, and let $g_\phi$ return standardized register
predictions. Define the encoder-coordinate predictor by
$f_\phi=\mathcal S_r^{-1}\circ g_\phi\circ\mathcal S_p$.
We minimize
\begin{equation}
\mathcal L_{\mathrm{reg}}(\phi)
=\mathbb E_{x\sim\mathcal D_{\mathrm{train}}}
\left[\operatorname{MSE}\!\left(
\mathcal S_r(f_\phi(Z_p(x))),\mathcal S_r(Z_r(x))
\right)\right].
\label{eq:register-prediction-loss}
\end{equation}
MSE is the unweighted mean over the minibatch, four registers, and
768 feature dimensions. Loss arithmetic is FP32; teacher and predictor
forwards use BF16, with TF32 disabled for the teacher and enabled for
the predictor.

\paragraph{Training and selection.}
Both encoders use the same optimization settings: AdamW with
learning rate $10^{-4}$,
$(\beta_1,\beta_2)=(0.9,0.95)$, weight decay $0.01$, and batch size 256,
retaining the final partial batch. We train for a fixed 100 epochs
(499,900 updates), with five epochs of linear warmup followed by cosine
decay to $10^{-5}$ and gradient-norm clipping at 1.0.
The 640 reserved calibration parents and content duplicates are excluded
from predictor fitting and normalization-statistic estimation.
Checkpoint selection minimizes standardized register MSE on 128 fixed,
training-derived selection images after each epoch; exact ties retain
the earlier epoch. Another 256 disjoint training-derived images provide
monitoring only. Final validation images are excluded from fitting and
checkpoint selection. The seed-0 checkpoints are selected at epoch 96
for DINOv2 and epoch 95 for DINOv3. All three DINOv2 decoder-seed
evaluations reuse the epoch-96 predictor and the same generated-patch bank.

\paragraph{Generation coordinate conversion.}
For the DINOv2 pretrained-XL experiment, source normalization of each
stored generated patch tensor is inverted exactly once. Raw patch
features are standardized for MHAP and independently converted to the
receiving RAE wrapper's patch coordinates. MHAP outputs return to
encoder coordinates through
$\hat r_{jc}=\mu_{r,jc}+\sigma_{r,jc}\hat r^{\mathrm{std}}_{jc}$.
The generation adapter then applies the receiver's auxiliary
normalization and calls \texttt{decode} with
\texttt{aux\_tokens\_are\_normalized=True}.
The RAE wrapper \emph{inverts} that normalization before passing the
registers to the learned decoder's projection and prepend interface.
Native reconstruction instead supplies raw register features with this
flag set to \texttt{False}. Thus both routes supply encoder-coordinate
register features to the learned decoder; generation's extra
normalize/inverse-normalize pair is an API coordinate conversion.
These wrapper transforms use $\sqrt{v+10^{-5}}$, distinct from the
predictor's standard-deviation floor. The wrapper likewise inverts its
patch normalization before decoding. No image, native-register target,
or CLS feature is available to the generation path.

\subsection{DINOv2 pretrained-XL evaluation}
\label{appendix:stage2-generation-setup}

The primary generation comparison uses a fixed bank of 50,000
DINOv2 patch representations sampled by the pretrained DiT-DH/XL.
The bank was generated using the Euler ODE sampler with
\texttt{num\_steps=50}, classifier-free guidance scale 1.0,
and global sampling seed 42. Sampling uses BF16 with TF32 enabled.
Every decoding route preserves the same stored patch tensors,
image identities, and class labels.

The stored patch bank uses FP16 tensors.
Source normalization is inverted once before conversion to the
receiving decoder's coordinates.
The evaluated decoding procedure uses BF16 with TF32 enabled
and microbatches of eight.
Outputs are clamped to $[0,1]$, multiplied by 255, and truncated
to uint8 before evaluation. All routes use the same pinned
Inception evaluator and real-image reference.

\begin{table}[t]
\centering
\small
\begin{tabular}{lcc}
\toprule
Decoder seed
& \shortstack{Patch-only ViT-B\\gFID $\downarrow$}
& \shortstack{MHAP + REG4 ViT-B\\gFID $\downarrow$}\\
\midrule
0 & 1.818943 & 1.656117\\
1 & 1.813372 & 1.637629\\
2 & 1.824930 & 1.630376\\
\midrule
Mean $\pm$ SD
& $1.819082\pm0.005780$
& $1.641374\pm0.013273$\\
\midrule
\multicolumn{2}{l}{Released ViT-XL decoder} & 1.557247\\
\bottomrule
\end{tabular}
\caption{\textbf{DINOv2 generation by decoder seed.}
All rows decode the same 50,000 pretrained-XL patch samples.
MHAP uses predictor seed 0, selected at epoch 96, for every register
decoder. Aggregate rows report mean $\pm$ sample standard deviation
across decoder seeds. The released decoder is a separate ViT-XL
reference.}
\label{tab:appendix-xl-per-seed}
\end{table}

The released RAE decoder has gFWD 0.71607 against a fixed set of
10,000 real RGB images. This FWD reference differs from the
ADM-reference evaluation used for gFID.
The two measurements therefore use different representations and
reference supports. Reconstruction FWD uses its own evaluation
cohort and is not numerically interchangeable with generation FWD.

All decoding routes receive the same 50,000 stored patch samples.
Their outputs are evaluated using identical image postprocessing,
the same Inception evaluator, and the same real-image reference.
Training details for the register-input decoders appear in
Appendix~\ref{appendix:feature-loss-details}. The released ViT-XL decoder
is larger than the ViT-B register decoder and uses its own training
recipe. 

\subsection{Joint generation of patches and auxiliary tokens}
\label{appendix:joint-auxiliary-generation}

We also examine a separate ImageNet-1K experiment that jointly
generates DINOv2 patches and five auxiliary tokens:
CLS and four registers.
This experiment uses DiT-S generators trained for 80 epochs,
rather than the pretrained XL generator used in the main
comparison. The generators use a 12-block, width-384 backbone
with six attention heads and a two-block, width-2048 DDT head
with 16 attention heads.

\paragraph{Joint latent modeling.}
The joint generator uses separate timestep and token-type
embeddings for the two streams. Patches follow a linear
transport path, while auxiliary tokens use a GVP path.
We compare auxiliary-loss weights
$\lambda_{\mathrm{aux}}\in\{0.05,0.25,0.50\}$.
The patch-only and joint generators use the same 80-epoch
training budget, effective batch size 1024 with four-step
gradient accumulation, and training seed 42.
Their optimizer is AdamW with initial learning rate
$2\times10^{-4}$, $\beta=(0.9,0.95)$, and zero weight decay.
The learning rate remains constant for the first 40 epochs
and then decays linearly toward $2\times10^{-5}$ at a configured
800-epoch endpoint; training stops at epoch 80.

\paragraph{Sampling and decoding.}
Each evaluation uses 50,000 class-balanced samples,
50-step Euler integration, CFG scale 1.0, sampling seed 42,
and EMA generator weights.
The ViT-B receivers are decoder-seed-0 models trained for
16 epochs with patch-latent noise $\tau=0.8$.
The patch-only receiver consumes patches; the auxiliary
receiver prepends CLS and four registers.
For auxiliary recovery, a separately trained MHAP predictor
uses five learned queries, two attention-pooling blocks,
hidden width 768, 12 heads, and MLP ratio 4.
It predicts CLS and four registers from generated patches;
it is distinct from the four-register predictor used with
the pretrained-XL samples. The five-token predictor uses
seed 0 and the retained best checkpoint from a 100-epoch
training run.

These scores belong to the separate DiT-S experiment and are
not directly comparable with the pretrained-XL evaluation.
The joint-generation rows at auxiliary-loss weights 0.05 and
0.25 record the ImageNet-1K validation reference; the reference
identity for weight 0.50 is unavailable.

\begin{table}[t]
\centering
\small
\begin{tabular}{lr}
\toprule
Generation and decoding route & gFID $\downarrow$\\
\midrule
Patch generation, patch-only decoding & 5.5981\\
Patch generation, MHAP auxiliary recovery & \textbf{5.4041}\\
Joint generation, $\lambda_{\mathrm{aux}}=0.05$ & 7.7431\\
Joint generation, $\lambda_{\mathrm{aux}}=0.25$ & 6.8716\\
Joint generation, $\lambda_{\mathrm{aux}}=0.50$
& $7.1076^{\dagger}$\\
\bottomrule
\end{tabular}
\caption{\textbf{Joint auxiliary generation.}
Separate DiT-S experiment with five auxiliary tokens
(CLS and four registers), $\tau=0.8$ receivers, and
50,000 evaluated images.
$\dagger$ The retained metric row does not identify its FID
reference; this value is not used for cross-route comparisons.}
\label{tab:joint-auxiliary-generation}
\end{table}

At auxiliary-loss weights 0.05 and 0.25, joint generation has
higher gFID than patch generation followed by MHAP recovery
under the recorded common reference. This compares separately
trained generators, not different auxiliary sources at fixed
sampled patches. The experiment concerns CLS plus four registers;
four-register-only joint generation and the causal effect of
the transport-path choice are not isolated.

\subsection{Qualitative model and sample identities}
\label{appendix:stage2-qualitative}

\paragraph{Composite reconstruction and generation examples.}
Figure~\ref{fig:reconstruction-generation}(a) shows reconstruction
evaluation IDs 026865 (dogsled), 012929 (Samoyed), and
048933 (seashore), from top to bottom.
Panel (b) shows generated sample IDs 005256 (night stage),
001171 (underwater shark), and 001530 (heron on water).

All displayed decoder outputs use DINOv2 features and seed-0
ViT-B decoders trained on ImageNet-1K, evaluated after 16 epochs.
The patch-only model is the reference checkpoint.
Reconstruction supplies four native encoder registers to the
register-input decoder; generation supplies four predictions
from the seed-0 MHAP checkpoint selected at epoch 96.

Reconstruction triplets share the same evaluated source crop.
Each generation pair uses the same stored pretrained-DiT-DH/XL
patch tensor and has no paired source image.
The reconstruction examples were selected post hoc from the
retained candidate gallery to illustrate appearance differences;
they do not estimate the average improvement.
Generation examples were drawn from the existing retained gallery.
Quantitative comparisons use their complete fixed evaluation sets.
All photographs retain their complete $256\times256$ pixels,
without additional cropping, color adjustment, sharpening,
or other enhancement.

\section{Appearance Analysis and Evaluation}
\label{appendix:appearance-analysis}
\subsection{DISTS texture and structure contributions}
\label{appendix:dists-components}

We decompose DISTS for the ImageNet-100 appearance experiment.
Each of three decoder seeds reconstructs the same 5,000 validation
images using patch-only (A) and register-input (B) decoding.
This gives 30,000 source--reconstruction evaluations and
15,000 paired A/B image--seed comparisons.

Let $S_{\mu,j}$ and $S_{\mathrm{cov},j}$ denote the implemented
feature-mean and feature-covariance similarities, respectively.
The weighted texture and structure contributions are
\begin{equation}
\begin{aligned}
W &= \sum_j(\alpha_j+\beta_j),\\
T &= \sum_j\frac{\alpha_j}{W}(1-S_{\mu,j}),\qquad
S = \sum_j\frac{\beta_j}{W}(1-S_{\mathrm{cov},j}).
\end{aligned}
\label{eq:dists-component-contributions}
\end{equation}
The sums include RGB and five frozen modified-VGG feature banks,
following DISTS~\citep{ding2021dists}. Both contributions retain
the same normalization $W$; neither is divided by its own
component weight mass.

\begin{table}[t]
\centering
\small
\setlength{\tabcolsep}{3pt}
\begin{tabular}{lrrrr}
\toprule
Contribution & Patch-only & Registers & Change & 95\% interval\\
\midrule
Texture
& 0.085537 & 0.079748 & -0.005789
& $[-0.006686,-0.004891]$\\
Structure
& 0.037023 & 0.035759 & -0.001263
& $[-0.001549,-0.000977]$\\
Total DISTS
& 0.122559 & 0.115507 & -0.007052
& $[-0.008227,-0.005877]$\\
\bottomrule
\end{tabular}
\caption{\textbf{DISTS component changes on ImageNet-100.}
Values average the three decoder-seed means. Changes are
register-input minus patch-only. Confidence intervals use paired
seed differences and Student's $t$ with two degrees of freedom.}
\label{tab:dists-components-completed}
\end{table}

Both contributions decrease in every decoder seed. Texture and
structure account for 82.1\% and 17.9\% of the weighted DISTS
reduction, respectively. These percentages partition the metric
change, not the amount of preserved texture or the causes of the
reconstruction improvement. The maximum per-image discrepancy
between DISTS and $T+S$ is $1.975\times10^{-7}$, and all 30,000
total scores reproduce the previously recorded DISTS values.

\subsection{Residual spectra}
\label{appendix:residual-spectra}

The Fourier analysis uses 30,000 A/B reconstructions:
5,000 images, two decoding routes, and three decoder seeds.
Smooth and detailed windows are selected from the original images
and used for both reconstructions.

Relative to patch-only decoding, register-input decoding reduces
residual-gradient MSE by $1.372\times10^{-5}$ in smooth windows,
with a 95\% seed interval of
$[1.120,1.624]\times10^{-5}$.
In detailed windows, the reduction is $9.117\times10^{-4}$,
with interval $[7.538,10.696]\times10^{-4}$.

Output gradient power also decreases in detailed windows.
Thus, lower residual-gradient error co-occurs with lower output
gradient power in detailed regions.
The measurements describe changes in fidelity to the original
gradient field alongside changes in the reconstructed field itself.
Radial averaging removes orientation information; this analysis
measures residual magnitude by radial frequency.

\subsection{Register substitutions}
\label{appendix:appearance-interventions}

These measurements use the ImageNet-100
\texttt{B\_preg\_standard} decoders, training seeds 0, 1, and 2,
on the same 5,000 validation images. They are distinct from the
ImageNet-1K experiments. The four-way comparison in
Equation~\ref{eq:appearance-four-way} crosses original and transformed
patch/register states within each fixed decoder; decoder weights,
register positions, normalization, and output procedure remain fixed.

\paragraph{Joint color and saturation transformation.}
For each spatial location of the original sRGB-coded image
$x\in[0,1]^{3\times H\times W}$, define
\begin{align}
g&=(1.08,0.98,0.90),\quad b=(0.015,0,-0.005),\\
y &= g\odot x+b,
\ell &= 0.2126y_R+0.7152y_G+0.0722y_B,\\
T(x)_c &= \operatorname{clip}_{[0,1]}
       \bigl(\ell+1.12(y_c-\ell)\bigr).
\end{align}
Thus the saturation pivot is the per-pixel luma of the
\emph{already gain-and-bias-transformed} RGB vector.
There is no separate luminance-contrast pivot, spatial resampling,
or conversion to linear-light RGB. Clipping occurs only after the
saturation operation; 4.1886\% of the channel values require clipping.
The transformed image remains floating point for encoding and readout.
All intervention metrics use decoder outputs clamped to $[0,1]$
\emph{before} rounding to the archived uint8 reconstructions.

\paragraph{Directional aggregation.}
Let $m(z)\in\mathbb R^3$ be the vector of spatial channel means,
$d_i=m(T(x_i))-m(x_i)$, and
\begin{align}
u^{P}_{s,i}
 &=m(D_s(P_i,R_{T,i}))-m(D_s(P_i,R_i)),\\
u^{P_T}_{s,i}
 &=m(D_s(P_{T,i},R_{T,i}))-m(D_s(P_{T,i},R_i)).
\end{align}
For either patch state $Q\in\{P,P_T\}$, the within-seed coefficient is
\begin{equation}
\beta_s^{Q}
 = \frac{\sum_{i\in\mathcal I}(u^Q_{s,i})^\top d_i}
         {\sum_{i\in\mathcal I}\|d_i\|_2^2},
\qquad
\mathcal I=\{i:\|d_i\|_2\geq10^{-3}\}.
\end{equation}
This is a pooled slope through the origin, equivalently a
$\|d_i\|_2^2$-weighted mean of the individual projection coefficients.
It is not the unweighted mean of per-image ratios.
Below the threshold, projection, cosine, numerator, and denominator
readouts are marked undefined and excluded from pooling;
no additive denominator epsilon defines a finite slope there.
All 5,000 RGB directions are defined in each of the four cells
for all three seeds.

Reported coefficients are the arithmetic means of the three
within-seed slopes. The original-patch and transformed-patch
coefficients are respectively $0.49943$ and $0.55917$;
the paired interaction is $0.05974$.
The four cells share the same directions, denominator, and support
within each seed, so the transformed-patch effect is the difference
of their pooled slopes. Decoder training seed is the outer unit:
95\% intervals use $\bar\beta\pm4.30265\,s_\beta/\sqrt{3}$,
with sample standard deviation $s_\beta$ across seeds.

\paragraph{Orthogonal response.}
For the register substitution at \emph{original} receiver patches,
the per-image orthogonal response is
\begin{equation}
o_{s,i}=\left\|u^P_{s,i}
 -\frac{(u^P_{s,i})^\top d_i}{\|d_i\|_2^2}d_i\right\|_2.
\end{equation}
Its reported value, $0.01552$ in $[0,1]$ RGB units, is the arithmetic
mean of these Euclidean norms over images within seed and then over
three seeds. It is not the norm of an averaged response or an RGB RMS.
For a direction below threshold, the implementation retains the full
change norm as a finite diagnostic by setting the subtracted component
to zero; no such cases occur in this appearance comparison.

\paragraph{Edge and semantic readouts.}
The edge statistic is cosine similarity between flattened
\emph{Sobel-magnitude} maps. For the sRGB-coded grayscale combination
$L(z)=0.2126z_R+0.7152z_G+0.0722z_B$, define
\begin{align}
K_x&=\begin{bmatrix}-1&0&1\\-2&0&2\\-1&0&1\end{bmatrix},\qquad K_y=K_x^\top,\\
e(z)&=\operatorname{vec}\sqrt{
 (K_x*L(z))^2+(K_y*L(z))^2+10^{-12}}.
\end{align}
Convolutions use one-pixel zero padding and the unnormalized kernels
shown above. Cosine similarity clamps each vector norm below at
$10^{-8}$. The reported change averages
\[
\cos(e(D_s(P_i,R_{T,i})),e(x_i))-
 \cos(e(D_s(P_i,R_i)),e(x_i))
\]
over images within seed and then over seeds, giving $+0.000374$.
This compares gradient magnitudes, not gradient orientations.

Under the same original-patch register substitution, supervised
ConvNeXt-Tiny logit cosine to the original image changes from
$0.92937$ to $0.92420$. Top-1 agreement changes by
$-0.0133$ percentage points; its 95\% seed interval crosses zero.
The evaluator uses the fixed ImageNet-1K weights, bilinear
$256\!\to\!236$ resizing, a centered $224\times224$ crop, and ImageNet
channel normalization. These results describe appearance responses and accompanying
fidelity/semantic readouts. Channel-wise color and saturation change
jointly in this intervention. Independent brightness, color-balance,
and contrast control requires separate interventions.

\subsection{Register interpolation with fixed receiver patches}
\label{appendix:register-interpolation}
\label{appendix:mixed-scene-register-interpolation}

\paragraph{Decoder and image pairs.}
The trajectories in Figure~\ref{fig:register-appearance-control}
use the DINOv2 four-register ViT-B decoder trained on ImageNet-1K
for 16 epochs, decoder seed 0. Each source image is encoded once.
During interpolation, receiver patches, decoder weights, and token
positions remain fixed; only the four register inputs change.
Neither encoder CLS nor MHAP predictions are supplied.

The displayed pairs are \texttt{p02\_ba} and \texttt{p10\_ab}.
For \texttt{p02\_ba}, the receiver is
\nolinkurl{Places365_val_00013976.jpg} and the donor is
\nolinkurl{Places365_val_00030622.jpg}.
For \texttt{p10\_ab}, the receiver is
\nolinkurl{ILSVRC2012_val_00031709.JPEG} and the donor is
\nolinkurl{ILSVRC2012_val_00006179.JPEG}.
These display pairs were selected after inspecting the saved
trajectories. The source-image experiment contains ten preselected
pairs evaluated in both directions, giving 20 retained trajectories.

\paragraph{Slotwise interpolation.}
For register slot $j\in\{1,\ldots,4\}$, let $r_j$ and $r_j^\star$
be the receiver and donor features in terminal encoder coordinates.
For nonzero endpoints, define
\[
u_j=\frac{r_j}{\|r_j\|_2},\qquad
v_j=\frac{r_j^\star}{\|r_j^\star\|_2},\qquad
\theta_j=\arccos(u_j^\top v_j).
\]
For $0<\theta_j<\pi$,
\begin{equation}
\begin{aligned}
r_j(\alpha)
&=\rho_j(\alpha)\left[
\frac{\sin((1-\alpha)\theta_j)}{\sin\theta_j}u_j
+\frac{\sin(\alpha\theta_j)}{\sin\theta_j}v_j
\right],\\
\rho_j(\alpha)
&=(1-\alpha)\|r_j\|_2+\alpha\|r_j^\star\|_2.
\end{aligned}
\label{eq:appendix-register-slerp}
\end{equation}
Directions follow the shortest spherical arc and norms are
interpolated linearly. All four slots use the same $\alpha$,
without slot mixing, centering, or whitening. The eight slot pairs
shown in Figure~\ref{fig:register-appearance-control} use this
non-degenerate spherical branch. At $\alpha=0$ and $\alpha=1$,
the original endpoint tensors are supplied directly.

We evaluate $\alpha\in\{0,0.125,\ldots,1\}$.
Interpolation arithmetic uses FP64; intermediate registers are
converted to FP32 before decoding. Inference uses CPU BF16
autocast with batch size one. Each curve contains the nine recorded
measurements for one pair, without averaging across receivers.
The image panels display $\alpha=0,0.25,0.5,0.75,1$.

\paragraph{Luminance and gradient-energy measurements.}
Let $x_\alpha=\operatorname{clip}_{[0,1]}(\hat x_\alpha)$ denote
the decoded RGB image, with $H=W=256$.
Measurements use these clipped floating-point outputs before
rounding to uint8 for display. Let
$w=(0.2126,0.7152,0.0722)^\top$ and define the channelwise
sRGB-to-linear conversion
\[
\ell(s)=
\begin{cases}
s/12.92, & s\leq0.04045,\\
((s+0.055)/1.055)^{2.4}, & s>0.04045.
\end{cases}
\]
Mean linear-light luminance is
\begin{equation}
\overline{Y}(x_\alpha)
=\frac{1}{HW}\sum_{h=1}^{H}\sum_{k=1}^{W}
w^\top\ell(x_{\alpha,hk}).
\label{eq:interpolation-luminance}
\end{equation}
Gradient energy instead uses the sRGB-coded luma
$q_{hk}=w^\top x_{\alpha,hk}$. With spatial mean $\bar q$, define
\begin{equation}
\begin{aligned}
E(q)
&=\frac{1}{H(W-1)}
\sum_{h=1}^{H}\sum_{k=1}^{W-1}(q_{h,k+1}-q_{hk})^2\\
&\quad+\frac{1}{(H-1)W}
\sum_{h=1}^{H-1}\sum_{k=1}^{W}(q_{h+1,k}-q_{hk})^2,\\
V(q)
&=\frac{1}{HW}\sum_{h=1}^{H}\sum_{k=1}^{W}(q_{hk}-\bar q)^2,
\qquad G(q)=\frac{E(q)}{V(q)}.
\end{aligned}
\label{eq:interpolation-gradient-energy}
\end{equation}
Horizontal and vertical differences are averaged separately over
their valid adjacent-pixel pairs, without boundary padding.
The ratio is undefined for $V(q)\leq10^{-12}$; all displayed
measurements have variance above this threshold.

\paragraph{All retained trajectories.}
We summarize all 20 directed trajectories from the ten selected
image pairs, using all nine interpolation values per trajectory.
For a statistic $m\in\{\overline Y,G\}$, let
$m_\alpha=m(D_R(P_r,R_\alpha))$ and define
\[
g_m=m(D_R(P_d,R_d))-m_0,\qquad
\Delta_m=m_1-m_0.
\]
Thus, donor direction is determined by the donor's versus
receiver's native reconstruction, not their source photographs.
For $|g_m|>10^{-12}$, set $s_m=\operatorname{sign}(g_m)$.
An endpoint is aligned when $s_m\Delta_m>10^{-12}$.
A donor-directed monotonic trajectory satisfies
\[
s_m(m_{(j+1)/8}-m_{j/8})\geq-10^{-12}
\quad\text{for all }j=0,\ldots,7.
\]

\begin{table}[t]
\centering
\small
\setlength{\tabcolsep}{4pt}
\begin{tabular}{llrrrr}
\toprule
Statistic & Donor & $n$ & Aligned & Monotonic
& Median $\Delta$\\
\midrule
Luminance & Lower  & 10 & 10 & 7 & -0.092725\\
Luminance & Higher & 10 &  9 & 7 & +0.049684\\
Normalized gradient energy & Lower
& 10 & 10 & 8 & -0.203004\\
Normalized gradient energy & Higher
& 10 & 10 & 9 & +0.065271\\
\bottomrule
\end{tabular}
\caption{\textbf{All retained register-interpolation trajectories.}
DINOv2 four-register decoder trained on ImageNet-1K, decoder
seed 0. Donor indicates a lower or higher native reconstructed
statistic. Aligned counts donor-directed endpoint changes;
monotonic counts trajectories following that direction across
all eight adjacent steps. Medians include adverse responses.}
\label{tab:all-register-trajectories}
\end{table}

Luminance follows the donor-defined direction at 19/20 endpoints,
with 14/20 donor-directed monotonic trajectories. Normalized
gradient energy follows that direction at 20/20 endpoints,
with 17/20 monotonic trajectories. The opposite luminance response,
\texttt{p08\_ab}, remains included. One gradient-energy response
overshoots the donor statistic. No neutral endpoints or
degenerate donor-direction differences occur.

These counts describe the retained image pairs. Interpolation
steps, reciprocal directions, and pairs sharing source images
are not independent experimental replicates; no population-level
confidence intervals are assigned. Directional alignment does
not require reaching the donor statistic or preserving every
other image property.

\subsection{Wavelet distance}
\label{appendix:appearance-metric-sensitivity}

FWD uses the authors' Haar level-four wavelet-packet
implementation~\citep{veeramacheneni2025fwd}.
For the three-seed ImageNet-100 appearance experiment, mean FWD
is 0.86126 for patch-only decoding and 0.83224 for register-input
decoding. The paired change is $-0.02902$, with an unadjusted
95\% seed-$t$ interval of $[-0.07072,0.01267]$.
The register-input model has lower FWD in each seed, but the
interval includes zero.

\clearpage
\section{Additional Reconstruction and Generation Examples}
\label{appendix:qualitative-reconstruction}

\begin{figure}[!htbp]
\centering
\includegraphics[
  width=\linewidth,
  height=0.7\textheight,
  keepaspectratio
]{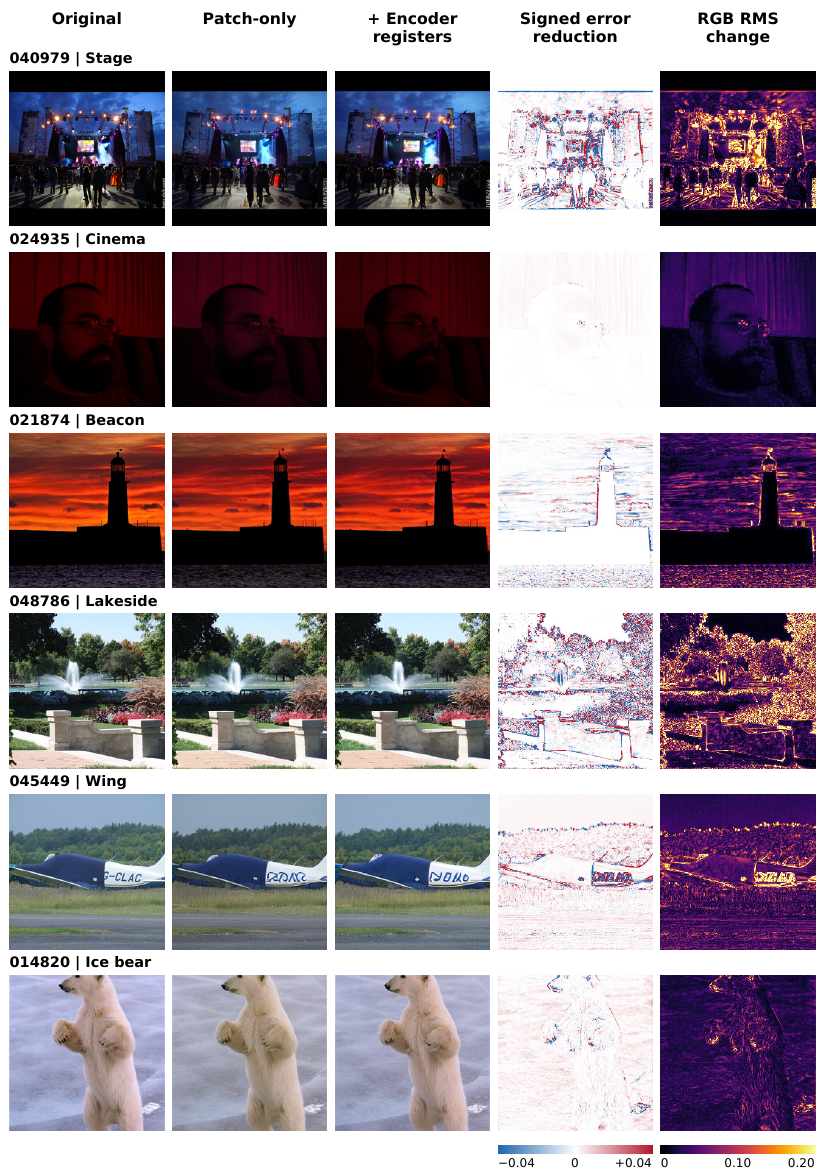}
\caption{\textbf{Additional ImageNet-1K reconstructions.}
Columns show the original image, patch-only reconstruction,
reconstruction with four encoder registers, signed RGB MSE
reduction, and RGB RMS change. Signed maps show patch-only minus
register-input squared error against the original, averaged over
RGB channels: red indicates lower error with registers and blue
lower error with patches alone. RMS maps show unsigned differences
between reconstructions. Fixed scales are $[-0.04,0.04]$ for signed
error reduction and $[0,0.20]$ for RMS change; values outside these
ranges saturate for display.}
\label{fig:appendix-in1k-reconstruction-maps}
\end{figure}

\begin{figure}[!htbp]
\centering
\includegraphics[
  width=\linewidth,
  height=0.76\textheight,
  keepaspectratio
]{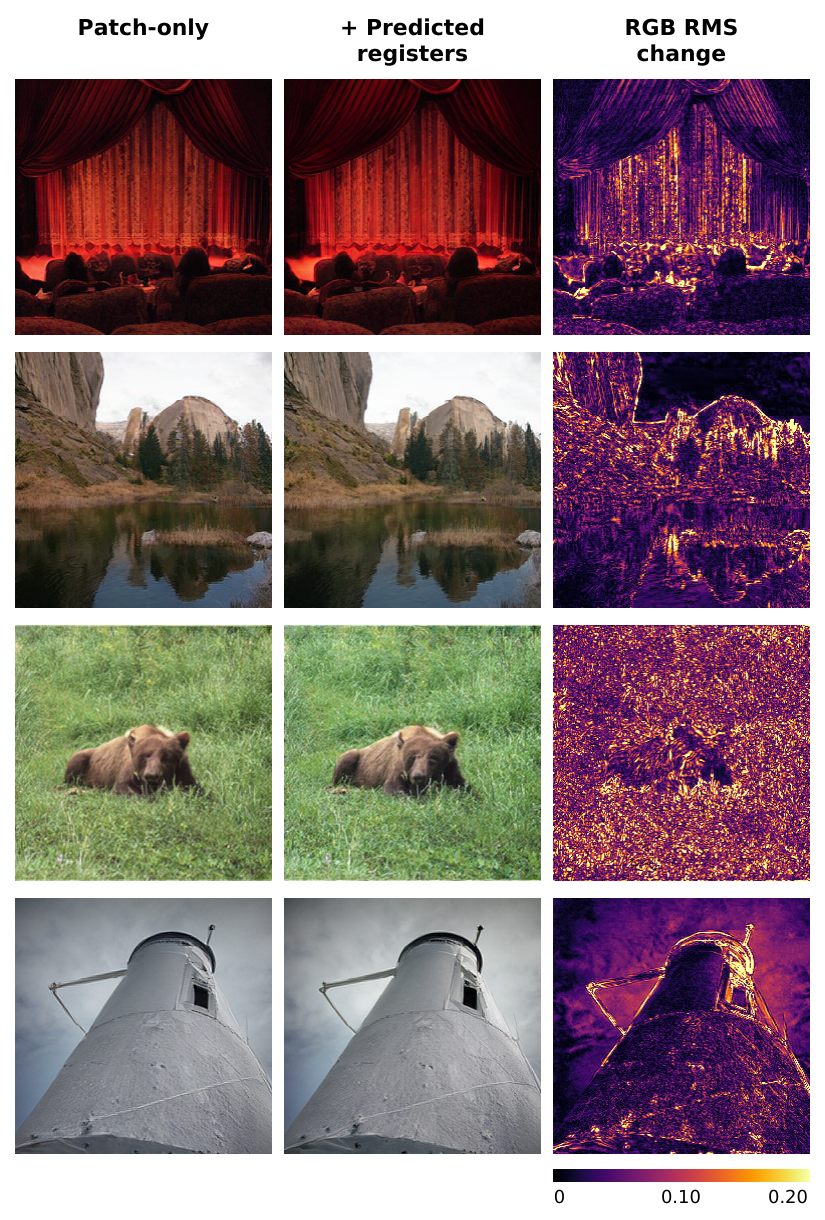}
\caption{\textbf{Additional generation examples.}
Each row decodes the same stored pretrained DiT-DH/XL patch tensor
using a patch-only decoder or a decoder supplied with
MHAP-predicted registers. The right column shows the per-pixel
RGB RMS difference between outputs on a fixed $[0,0.20]$ scale.
Brighter values indicate larger output changes; values above
0.20 saturate for display. These maps measure change magnitude,
not a difference in generation quality.}
\label{fig:appendix-generation-rms}
\end{figure}

Figures~\ref{fig:appendix-in1k-reconstruction-maps}
and~\ref{fig:appendix-generation-rms} show six reconstruction
examples and four generated-image pairs, respectively. These
selections use distinct classes within each gallery and exclude
image identities shown in Figure~\ref{fig:reconstruction-generation}.
All displayed images use ImageNet-1K-trained DINOv2 ViT-B decoders;
model and sample identities are given in
Appendix~\ref{appendix:stage2-qualitative}.

\clearpage

\end{document}